\pdfoutput=1

\documentclass[11pt]{article}
\PassOptionsToPackage{dvipsnames,usenames}{xcolor}

\usepackage[final]{acl}

\usepackage{times}
\usepackage{latexsym}

\usepackage[T1]{fontenc}

\usepackage[utf8]{inputenc}

\usepackage{microtype}

\usepackage{inconsolata}

\usepackage{graphicx}
\usepackage{amsmath}
\usepackage{amssymb}
\usepackage{cleveref}
\usepackage{multirow}
\usepackage{graphicx}
\usepackage{array}
\usepackage{booktabs}
\usepackage{caption} 
\usepackage{siunitx} 
 \usepackage{xspace}
\usepackage{xcolor}
\usepackage{svg}
\usepackage{siunitx}
\usepackage{bbm}

\newcommand{\spv}{\textsc{\textbf{spv}}\xspace}
\newcommand{\foi}{\textsc{\textbf{foi}}\xspace}
\newcommand{\hcs}{\textsc{\textbf{hcs}}\xspace}

\newcommand{\prompt}[1]{{\textcolor{Black}{#1}}}
\newcommand{\llmError}[1]{\textcolor{Black}{\textbf{#1}}}
\newcommand{\predictionstyle}[1]{\textcolor{Black}{\ensuremath{\hat{#1}}}}
\newcommand{\myGranularity}[1]{{\textcolor{Black}{#1}}}

\newcommand{\metricsStyle}[1]{\textcolor{Black}{#1}}

\newcommand{\feat}{\textcolor{Black}{f}}
\newcommand{\featSet}{\textcolor{Black}{\mathcal{F}}}

\newcommand{\myprompt}[2]{\newcommand{#1}{\prompt{#2}}}
\newcommand{\promptError}[2]{\newcommand{#1}{\llmError{#2}}}
\newcommand{\llmpred}[2]{\newcommand{#1}{\predictionstyle{#2}}}
\newcommand{\granularity}[2]{\newcommand{#1}{\myGranularity{#2}}}
\newcommand{\mymetrics}[2]{\newcommand{#1}{\metricsStyle{#2}}}
\newcommand{\ginG}{\ensuremath{\myGranularity{g \in G}}}
\newcommand{\metricinMetrics}{\ensuremath{\metricsStyle{ k\in \{\mathrm{Surp}, \mathrm{CIS}, \mathrm{CWS}, \mathrm{Entropy}\}}}}

\newcommand{\Sgranular}[1]{\Ssymbol^{\myGranularity{#1}}}

\newcommand{\errorLabelFull}{\textcolor{Black}{\textbf{\Esymbol{} $\in \{0, 1\}$}}}

\myprompt{\Ssymbol}{S_j}
\myprompt{\Sjexpr}{S_{j,\mathrm{expr}}}

\myprompt{\Isymbol}{I}
\myprompt{\Qsymbol}{Q}

\promptError{\Esymbol}{E}
\llmpred{\ypred}{y_j}
\mymetrics{\metric}{\Phi\xspace}
\granularity{\gsymbol}{g}
\granularity{\Gsymbol}{G}

\title{From Input Perception to Predictive Insight: Modeling Model Blind Spots Before They Become Errors}

\author{
  Maggie Mi\textsuperscript{1} \quad
  Aline Villavicencio\textsuperscript{2,3,4} \quad
  Nafise Sadat Moosavi\textsuperscript{1} \\
  \textsuperscript{1}University of Sheffield \quad
  \textsuperscript{2}University of Exeter \quad
  \textsuperscript{3}The Alan Turing Institute \quad
  \textsuperscript{4}UFRN, Brazil \\
    \texttt{\{zmi1, n.s.moosavi\}@sheffield.ac.uk} \\
  \texttt{a.villavicencio@exeter.ac.uk}
}

\begin{document}
\maketitle
\begin{abstract}
Language models often struggle with idiomatic, figurative, or context-sensitive inputs, not because they produce flawed outputs, but because they misinterpret the input from the outset. We propose an input-only method for anticipating such failures using token-level likelihood features inspired by surprisal and the Uniform Information Density hypothesis. These features capture localized uncertainty in input comprehension and outperform standard baselines across five linguistically challenging datasets. We show that span-localized features improve error detection for larger models, while smaller models benefit from global patterns. Our method requires no access to outputs or hidden activations, offering a lightweight and generalizable approach to pre-generation error prediction. 

\hspace{.5em}\includegraphics[width=1.25em,height=1.25em]{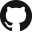}{\hspace{.75em}\parbox{\dimexpr\linewidth-2\fboxsep-2\fboxrule}{\footnotesize \url{https://github.com/mi-m1/input_perception}}}

\end{abstract}

\section{Introduction}
Model failures in language understanding do not always stem from incoherent outputs. Instead, they may originate earlier in the processing pipeline, from how the model internally interprets the input. When meaning depends on context-sensitive constructions, such as idiomatic or metaphorical expressions, the model may produce a plausible response that is nevertheless grounded in a misreading of the input. These cases reveal a blind spot in the model’s comprehension, where high-confidence generation masks a fundamental interpretive error.
This raises a critical question: \textit{Can we anticipate such errors before the model generates a response, purely by examining how it internally processes the input?} Prior work has shown that models tend to perform better on inputs they assign higher overall likelihoods \citep{ohi-etal-2024-likelihood, mccoy2024embers}, suggesting that token-level probabilities may encode latent signals of confidence or uncertainty. However, most existing approaches reduce this likelihood information to a global scalar, such as perplexity, and do not examine how fine-grained variations across the input might reflect deeper patterns of misalignment. Moreover, nearly all existing techniques for error or uncertainty estimation rely on decoding-time cues, such as logits \cite{tuned-lens}, entropy \cite{pereyra2017regularizingneuralnetworkspenalizing}, or sampling variance. Our method, however, is complementary: it anticipates failure at the input stage without consulting the output.


\begin{figure*}[t]
    \centering
    \includegraphics[width=0.9\textwidth]{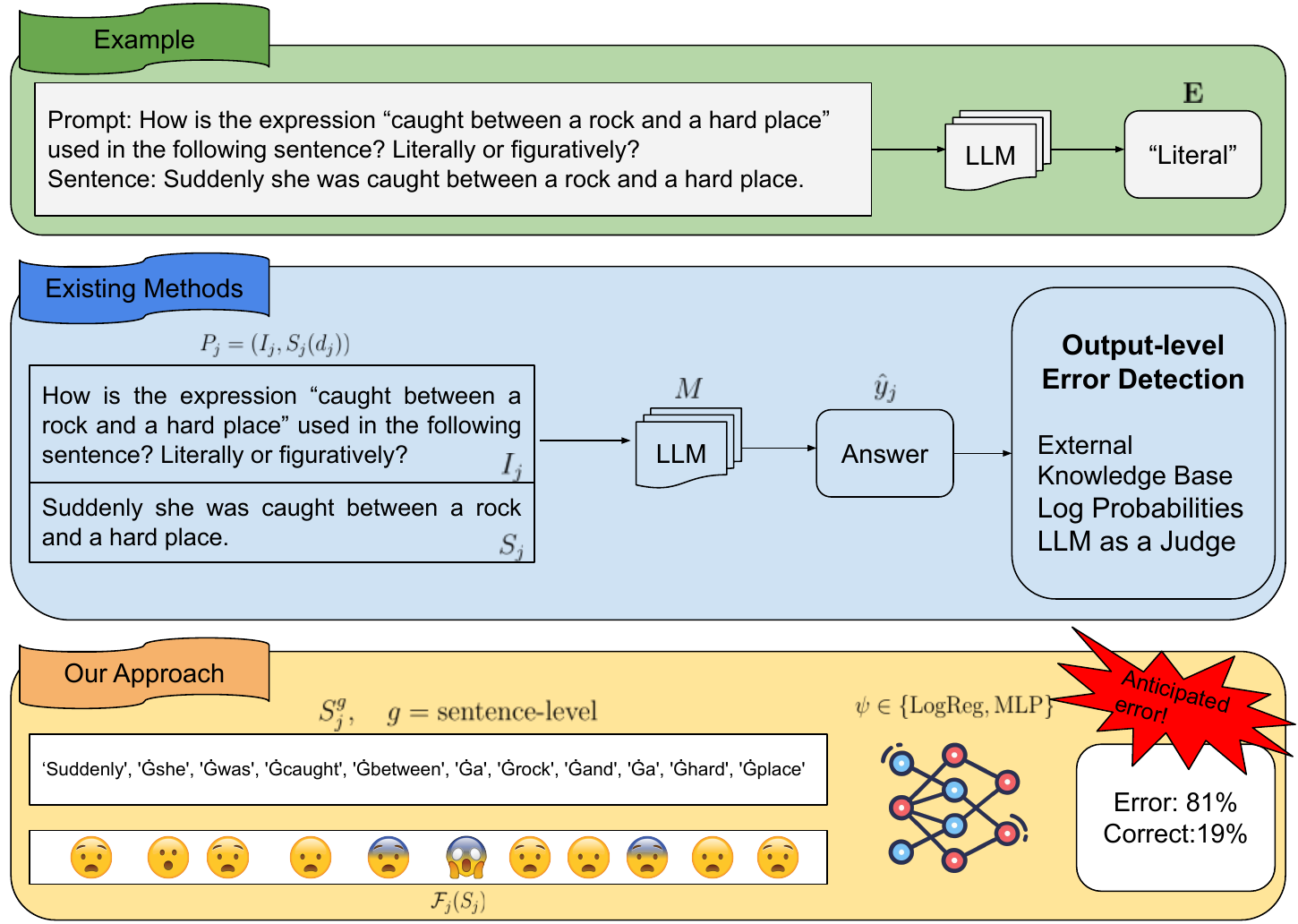}
    \caption{An illustration of an LLM failure on a non-compositional semantics task (top panel). Standard error detection methods typically work by checking the generated output afterward, often requiring extra resources such as a Judge LLM (middle panel). In contrast, our method applies information-theoretic measures directly to the input sentence (bottom panel). In this case, our model estimated an 81\% likelihood that the LLM would fail on this example, and indeed, it did.}
    \vspace{-5pt}
    \label{fig:overview}
\end{figure*}

We propose an input-driven framework for anticipating language model (LM) errors by analyzing the structure of the likelihood surface over input sequences. Our approach is motivated by the Uniform Information Density (UID) hypothesis \citep{jaegerLevy_2006, jaeger2010redundancy}, which views language as a signal optimized to distribute information evenly across an utterance.


In practice, information density fluctuates due to grammatical, discourse, and pragmatic factors \citep{levy2013, genzel-charniak-2002-entropy, xu-reitter-2016-entropy}, and forms information contours that reflects meaningful variations of contextual information  \citep{tsipidi-etal-2024-surprise}. Similarly, non-literal language, e.g., idioms, metaphors, metonymy, violates compositional expectations and disrupts local information uniformity. For LMs, such irregularities appear as perturbations in the likelihood landscape. We hypothesize that these points of instability, where predictive expectations diverge from natural variability, offer useful signals for anticipating model errors.

Although our framework applies to many kinds of context-sensitive meaning, we target idioms, metaphors, and metonymy because they stress contextual interpretation and have been repeatedly shown be be challenging for LLMs: misunderstandings of context can flip the label from literal to figurative (or vice versa) and yield large errors. Recent studies show that even strong models often fail to leverage context for these distinctions \cite{mi-etal-2025-rolling,phelps-etal-2024-sign}, making these tasks an incisive testbed for input-side error prediction. A further advantage is observability: these datasets come with explicit span annotations for the potentially problematic phrases (e.g., the idiom or the metaphoric span). This lets us localize input-likelihood features to known points of interpretive risk and investigate whether the model’s own input-side signals anticipate mistakes more accurately than coarse global heuristics.



Our results show that token-level likelihood features, without access to output logits or hidden states, significantly improve error detection, particularly for smaller models, and outperform established baselines such as log probability, max token confidence, or Oddballness.

While our span-localized features leverage task-informed linguistic structure, our sentence-level features generalize across settings and suggest a broader insight: that the internal likelihood surface over the input encodes rich, interpretable signals of model comprehension. This opens the door to future work on black-box risk estimation that operates before generation, using only how the model ``reads'' the input as a basis for identifying when it is likely to fail.

\paragraph{Contributions}
(1) We propose an input-driven framework for anticipating language model errors using token-level likelihood features, without relying on outputs or internal activations.
(2) We develop both global and span-localized uncertainty features; while the latter are guided by task-specific linguistic structure, the global features are broadly applicable and offer a path toward dynamic localisation.
(3) We demonstrate the effectiveness of this approach across five linguistically grounded datasets, showing substantial gains over standard input-likelihood heuristics, particularly for smaller models (1B-3B parameters).

\section{Related Work}
\label{sec:related}

\paragraph{Literal vs. Figurative Understanding}
The task of detecting metaphors, idioms, and metonymy involves determining whether a linguistic expression is used figuratively or literally. Despite substantial progress, this remains a challenging problem for language models \cite{phelps-etal-2024-sign, he-etal-2024-enhancing, yang-etal-2024-chatgpts, tian-etal-2024-bridging, steen}. A key difficulty lies in models' inability to effectively leverage context for disambiguation. For example, \citet{mi-etal-2025-rolling} show that even under controlled, contrastive evaluation settings, large language models struggle to use contextual cues to distinguish figurative from literal interpretations. 

\citet{kabbara-cheung-2022-investigating} demonstrate that Transformer-based models often rely on superficial lexical or structural cues, rather than engaging in deeper pragmatic reasoning. This finding aligns with our hypothesis that many model errors arise not from output generation, but from internal misinterpretation of the input. If models rely on superficial cues rather than deeper semantic reasoning, then fine-grained input likelihood patterns, such as local surprisal spikes, may offer a more faithful signal of confusion or misalignment. By analyzing these token-level signals, our approach seeks to uncover where and how models exhibit shallow comprehension, especially in linguistically complex regions.




\paragraph{Surprisal and Psycholinguistics Signals}

Surprisal, rooted in Shannon’s information theory \citep{shannon-1948}, measures the unexpectedness of a word in context, with less predictable words imposing greater processing difficulty on humans \citep{hale-2001-probabilistic, levy-2008}. A large of body of literature has shown that surprisal values estimated from neural LMs (e.g., LSTMs and Transformers) predict human reading times \cite{oh-schuler-2023-transformer,wilcox_11_languages,pimentel2013_readingTimes,rambelli-etal-2023-frequent, goodkind-bicknell-2018-predictive}.

Other contextual predictors have also been explored. Entropy captures uncertainty about upcoming words, reflecting the number of plausible continuations in context \citep{futrell_lossyContext_surprisal}. Pointwise Mutual Information (PMI) \citep{fano_pmi, church-hanks-1990-word} quantifies word associations, and has long been used in modeling lexical co-occurrence and semantic similarity. Together, these measures capture token-level dimensions of processing difficulty: unexpectedness (surprisal), contextual uncertainty (entropy), and pairwise association strength (PMI). While primarily applied to model human comprehension, recent work reframes these signals as useful diagnostics for LMs themselves \citep{opedal-etal-2024-role}.

\paragraph{Error detection} 

A broad body of work has explored error detection and uncertainty in language models, often focusing on output-driven signals. Selective prediction \citep{gu-hopkins-2023-evaluation}, entropy-based methods \citep{shorinwa2024survey,wu-etal-2025-improve-decoding}, and token-level uncertainty estimation \citep{ma2025estimating} estimate uncertainty during or after generation, by relying on output probabilities or decoding dynamics \cite{wu-etal-2025-improve-decoding}. Other approaches probe model internals, e.g., hidden states \cite{burns2024discoveringlatentknowledgelanguage, azaria-mitchell-2023-internal, ITI-li, zou2025representationengineeringtopdownapproach}, attention patterns \cite{chuang-etal-2024-lookback}, or gradients, to diagnose understanding \citep{ashok2025language,vig-belinkov-2019-analyzing,Chefer_2021_CVPR}. While effective, these methods require access to internal representations or model outputs, limiting their applicability to open-source or non-black-box settings.

In contrast, our approach estimates error risk before any generation occurs, using only token-level input likelihoods. This purely input-driven method allows us to infer model comprehension externally, making it compatible with black-box or API-only models. Our work is also motivated by studies showing that models perform better on inputs with higher overall likelihoods \citep{ohi-etal-2024-likelihood,mccoy2024embers}, and by techniques like Oddballness \citep{gralinski-etal-2025-oddballness}, which flag anomalous inputs via likelihood deviations. These findings suggest that the input likelihood surface carries important signals of conceptual instability, particularly in linguistically complex regions and thus, offers a lightweight and generalizable path for proactive error detection.

\vspace{-4pt}
\section{Method}
\label{sec:method}

\subsection{Modeling Comprehension through Input Likelihood Structure}
Our approach is based on the hypothesis that the likelihoods a language model assigns to tokens within an input sequence encode latent signals of its internal comprehension. Specifically, we posit that surprisal and related features derived from input token probabilities can reveal regions of uncertainty or misalignment that precede downstream errors. This view aligns with psycholinguistic findings showing that humans experience increased processing difficulty at points of high surprisal \citep{hale-2001-probabilistic, smith2013effect}, and recent evidence suggesting similar interpretive dynamics in LLMs \citep{ohi-etal-2024-likelihood}.

Rather than analyzing outputs or decoder activations, we focus exclusively on the model's perception of the input. Given a dataset $\mathcal{D}$ comprising instances $d_j \in \mathcal{D}$, and we construct a prompt, $P_j$, composed of two components: 
\begin{equation}
    \mathcal{P}_j = (\Qsymbol_j, \Ssymbol(d_j))
\end{equation}
where $\Qsymbol_j$ denotes the task instruction specifying the phenomenon relevant to task $j$ and $\Ssymbol$ denotes the contextual sentence for instance $d_j$.

We extract features from the likelihood distribution over $S_j$, treating this distribution as a proxy for the model’s internal interpretation of the input.\footnote{While we do not explicitly design or optimize prompts, we acknowledge that the distribution over $S_j$ is shaped by $I_j$ due to the model’s autoregressive architecture.}
These features are computed either over the entire sentence (global) or over spans informed by linguistic structure (localized).

\subsection{Measures}
\label{sec:measures}

We define a set of information-theoretic measures, as:
\[
\Phi = \{\mathrm{SPR}, \mathrm{H}, \mathrm{CWS}, \mathrm{CIS}\}
\]

Each measure, \( k \in \Phi \), captures a distinct facet of model's predicative behaviour over various granularity (e.g., $\Ssymbol$) of the input. These features require no task-specific information and are broadly applicable across inputs.

From these metrics, we derive sentence-level features by aggregating values across the context sentence ($\Ssymbol$) (e.g., mean surprisal, maximum entropy). Together, these features provide complementary perspectives on the model’s internal belief state, capturing token-level fit, uncertainty sharpness, confidence calibration, and contextual salience, and serve as interpretable signals for anticipating downstream error.

\paragraph{Surprisal}  
Surprisal measures the unexpectedness of a token \( t_i \), defined as the negative log-probability of the token given its preceding context:

\[
\text{SPR}(t_i) = -\log_2 P(t_i \mid t_{<i})
\]

\paragraph{Entropy.}  
Shannon entropy reflects the model’s uncertainty over the next token distribution at each position. Unlike surprisal, which is token-specific, entropy quantifies the overall spread of the model’s prediction:

\[
H(t_i) = -\sum_{w \in V} P(w \mid t_{<i}) \log_2 P(w \mid t_{<i})
\]

\paragraph{Confidence-Weighted Surprisal (CWS)}  
We propose a variant of surprisal that incorporates a penalty for low-confidence or diffuse next-token distributions. CWS augments surprisal with a KL divergence term measuring deviation from an idealized, peaked distribution:

\[
\text{CWS}(t_i) = -\log_2 P(t_i \mid t_{<i}) + \gamma \cdot D_{\text{KL}}(P \parallel Q)
\]

Here, \( P \) is the model’s predicted distribution, and \( Q \) is a reference distribution that assigns 0.9 probability to the observed token and distributes the remaining 0.1 uniformly. The penalty weight \( \gamma \) controls sensitivity to this divergence:
\[
D_{\text{KL}}(P \parallel Q) = \sum_{t \in V} P(t) \left[ \log_2 P(t) - \log_2 Q(t) \right]
\]

We introduce CWS to explicitly combine token correctness and sharpness of the belief, since there are settings where a model assigns moderately high probability to the observed input tokens but with a diffuse overall distribution (high entropy), thus, indicating low semantic commitment. This hybrid aims to penalise such diffuse predictions more directly.

\paragraph{Contextual Influence Score (CIS).}  
CIS measures the incremental informatoin that  \(t_i\) contributes to predicting \(t_{i+1}\):

\begin{align*}
\text{CIS}(t_i) =\ 
& \log_2 P(t_{i+1} \mid t_{\le_i}) \\
& - \log_2 P(t_{i+1} \mid t_{<i})
\end{align*}

Equivalently, it is a conditional PMI term. Operationally, the second term is computed by
rescoring \(t_{i+1}\) under the shortened prefix \(t_{<i}\) (i.e.,``without \(t_i\)'').
Thus, CIS measures the conditional PMI between  \( t_i \) and 
\( t_{i+1} \) given the prefix, and quantifies the incremental predictive contribution of  \( t_i \) during autoregressive inference.

\subsection{Features Localized to Challenging Input Regions}
\label{sec:localized-features}

While sentence-level features summarize the model’s input-conditioned belief profile, many reasoning failures stem from localized comprehension difficulties. In tasks with semantically complex constructions, the model must integrate context and resolve ambiguity within a confined span (e.g., the idiomatic phrase in $\Ssymbol$: $\Sjexpr$). We hypothesize that token-level likelihood patterns in such spans can more precisely indicate potential errors than global averages. When the boundaries of these challenging regions are known or can be inferred (e.g., via dataset annotation), we compute localized features informed by linguistic theory.
We operationalize three linguistically motivated hypotheses that link likelihood dynamics to characteristic comprehension challenges.

\paragraph{Fixedness of Idioms (\foi).}  
Idiomatic expressions are often syntactically rigid and lexically constrained \citep{chafe1968idiomaticity, fraser1970idioms}. Once the idiom is initiated, its subsequent tokens become increasingly predictable. In surprisal terms, we expect a decreasing pattern. We capture this with two features:

\textbf{Monotonic Decrease:} A binary feature indicating whether information decrease throughout the span:
\begin{align*}
\text{Decreasing}_{\text{expr}} = 
\mathbbm{1}\Big(& \forall\, i < j \in \Sjexpr, \\
& k
(w_i) > k(w_j) \Big)
\end{align*}
  
\textbf{Surprisal Spikes:} The number of local surprisal maxima in the span, reflecting unpredictability:

\begin{align*}
N_{\text{spikes}} = \Big| \big\{\, i \in \Sjexpr \,\big|\,
& k(w_i) > k(w_{i-1})\ \land \\
& k(w_i) > k(w_{i+1}) \,\big\} \Big|
\end{align*}

\paragraph{Selectional Preference Violations (\spv).}  
Metaphorical constructions often involve semantic mismatches between verbs and their arguments \citep{WILKS1975, WILKS1978}, which may lead to sharp information transitions. We define \textbf{Boundary Shift}, which represents the change in information from the final token in the span to the token immediately following it:
\[
\Delta_{\text{boundary}} = k(w_{\text{post}}) - k(w_{\text{end}})
\]
where \( w_{\text{end}} \) is the last token of the span, and \( w_{\text{post}} \) is the following token.

\paragraph{High Context Information (\hcs).}  
Comprehension failures often correlate with local information peaks. We test whether the model’s highest uncertainty is localized within the span using \textbf{Peak-in-Span Indicator:}
\[
\delta(x) = \mathbbm{1}(i^* \in \Sjexpr), \quad i^* = \arg\max_i k(w_i)
\]

We also design additional features that are inspired by there linguistic theories (see \Cref{sec:additional_features}. Collectively, these localized features are designed to complement sentence-level likelihood summaries by providing finer-grained signals of conceptual instability within semantically challenging parts of the input.

\paragraph{Granularities.} To capture cues for distinguishing literal from figurative usage, we compute features at four granularities. Sentence-level applies metrics to the full sentence. Expression-level restricts to the idiomatic span. Boundary-level focuses on words immediately flanking the idiom. Context-level uses the surrounding sentence with the idiom removed. These complementary views isolate internal, local, and contextual signals shaping interpretation.

\subsection{Feature Set}

For each tokenwise metric defined in \Cref{sec:measures} (e.g., surprisal, entropy, CIS, CWS),
and for each example $j$ with prompt $\mathcal{P}_j$ tokenized as
$\mathbf{x}_j=(x_{j,1},\dots,x_{j,m_j})$, we derive features at different
levels of {granularity}. With $\Phi$ denoting the set of measures and
$\mathcal{G}=\{$\textit{sentence-level},\ \textit{expression-level},\ \textit{boundary-level},\ \textit{context-level}\}
the granularities. For any region $g\in\mathcal{G}$, let $I_j^{g}\subseteq\{1,\dots,m_j\}$ be
the corresponding index set of tokens (e.g., all tokens for \textit{sentence-level},
the annotated span for \textit{expression-level}, etc.). We aggregate over a set of
operators $\mathcal{A}=\{\mathrm{mean},\mathrm{max}\mathrm{min}, \mathrm{std}\}$:

\[
\feat_{\prompt{j},\metricsStyle{k},\myGranularity{g},a}=
\begin{cases}
\dfrac{1}{|I_j^{g}|}\!\sum_{t\in I_j^{g}}\!\metricsStyle{k}(x_{j,1:t}), a=\text{mean},\\
a_{t\in I_j^{g}} \!\metricsStyle{k}(x_{j,1:t}),  a\in\{\text{max},\text{min},\text{std}\}.
\end{cases}
\]

Thus, our feature set for our linguistically targeted setting is:
\begin{align}
\featSet_{j}
= \big\{\, \feat_{\prompt{j},\,\metricsStyle{k},\,\myGranularity{g},\,a}
\;\big|\; \metricsStyle{k}\in\Phi,\ \myGranularity{g}\in\mathcal{G},\ a\in\mathcal{A} \,\big\}.
\end{align}

\paragraph{Sentence-level restriction.}
When restricting to sentence-level evaluations, we fix the granularity to sentence-level and apply the mean and maximum aggregations
across the sentence. Formally,
\begin{align}
\featSet^{\text{sent}}_{j}
= \big\{\, 
\feat_{\prompt{j},\,\metricsStyle{k},\,S_j,\,\mathrm{mean}},\ 
\feat_{\prompt{j},\,\metricsStyle{k},\,S_j,\,\mathrm{max}}
\;\big|\; \nonumber \\
\metricsStyle{k}\in\Phi
\,\big\}.
\end{align}

\vspace{-12pt}
\section{Experimental Setup}
\label{sec:evaluation}

\subsection{Datasets}

We utilize five benchmark datasets spanning three distinct types of non-literal language understanding: idiomaticity, metaphor, and metonymy. 

DICE \cite{mi-etal-2025-rolling} is an idiomaticity detection dataset that preserves the fixed lexical and syntactic form of idioms across both literal and figurative uses. Unlike prior datasets that alter idiom structure in literal cases, DICE forces models to rely on context rather than form to disambiguate meaning. MOH-X \cite{mohammad-etal-2016-metaphor} and TRoFi \cite{birke-sarkar-2006-clustering} are datasets for verbal metaphorical/literal identification. We use Task 14 \textit{Reference via Metonymy} of PUB \cite{sravanthi-etal-2024-pub} (henceforth, PUB 14), which are challenging cases where named entities refer not to themselves but to entities closely associated with them (e.g., "Washington" referring to the U.S. government). Similarly, ConMeC \cite{ghosh-jiang-2025-conmec} focuses on metonymy of common, high-frequency nouns (such as `glass' for `wine'). 

\subsection{Evaluation Paradigm}
\label{sec:prompting}

Given a LLM $M$, we use zero-shot prompts to assess the knowledge that the model has learned during training, rather than its ability to adapt to the task using in-context learning. Since understanding context is an inherent aspect of language ability, and not a downstream task, we aim to evaluate the model in its unaltered state. The only variation in the task instruction is the specification of the particular phenomenon relevant to each task. We provide our prompts for each task in \Cref{sec:task_prompts}. Thus, the model prediction is then given by:
\begin{equation}
    \ypred = M(\mathcal{P}_j)
\end{equation}

The accuracy of the model for each instance is evaluated by comparing \ypred with the gold label $y_j$. We provide the accuracy of the models evaluated in \Cref{sec:downstream_accuracy}.
\begin{equation}
    \text{Accuracy} = \frac{1}{|\mathcal{D}|} \sum_{j=1}^{|\mathcal{D}|} \mathbb{I}[\ypred = y_j]
\end{equation}

Let \(\mathbf{e}=(e_1,\ldots,e_{|\mathcal{D}|})\) denote the vector of error labels, where \(e_j = \mathbf{1}[\hat{y}_j \ne y_j]\).

\subsection{Classifiers}
We fit logistic regression and an MLP to map $\mathcal{F}_j$ to an error probability $\hat p_j$. For binary decisions we use a threshold $\tau$ (Appendix~B.3). Further details are in Appendix~B.3.

\subsection{Baselines}
Typically, methods such as log probability and max token probability have been used as signals for error detection. Thus, we fit our logistic regression model and MLP model on: log probability, max token probability and oddballness as baselines (see \Cref{sec:baselines_definitions}).

\begin{table*}[h]
\begin{center}
\small
\resizebox{\linewidth}{!}{%
\begin{tabular}{l *{5}{S} | *{5}{S}}
\toprule
Tasks & {DICE} & {MOHX} & {TroFi} & {PUB 14} & {ConMeC} & {DICE} & {MOHX} & {TroFi} & {PUB 14} & {ConMeC} \\ \midrule

\textbf{Baseline} & \multicolumn{10}{c}{\textsc{Max Probability Confidence}} \\ \cmidrule{2-11}
Llama-3.1-8B-Instruct &0 &0 &0 &74.8299319727891 &28.9017341040462 &0 &0 &0 &68.5714285714285 &0 \\
Llama-3.2-3B-Instruct &60.8187134502923 &0 &0 &95.4545454545454 &72.2415795586527 &68.1592039800995 &0 &0 &95.4545454545454 &72.2415795586527 \\
Llama-3.2-1B-Instruct &85.3185595567867 &76.6355140186915 &89.1963109354413 &85 &98.8970588235294 &84.958217270195 &76.6355140186915 &89.1963109354413 &85 &98.8970588235294 \\
Qwen2-1.5B-Instruct &79.5107033639143 &69.0721649484536 &74.9442379182156 &96.045197740113 &69.0734055354994 &76.8211920529801 &60.919540229885 &74.9442379182156 &96.045197740113 &70.4358068315665 \\
Qwen2.5-0.5B-Instruct &70.5882352941176 &63.8297872340425 &82.7057182705718 &98.342541436464 &89.4472361809045 &70.8097928436911 &67.032967032967 &82.7057182705718 &98.342541436464 &89.4472361809045 \\
Qwen2.5-7B-Instruct-1M &0 &0 &1.35593220338983 &97.2067039106145 &0 &0 &0 &0 &97.2067039106145 &0 \\
Qwen2.5-14B-Instruct-1M &0 &0 &0 &87.8048780487805 &0 &0 &0 &0 &87.8048780487805 &0 \\

\midrule
\textbf{Baseline} & \multicolumn{10}{c}{\textsc{Log Probability}} \\ \cmidrule{2-11}
Llama-3.1-8B-Instruct &22.754491017964 &0 &0 &74.8299319727891 &8 &0 &0 &0 &74.8299319727891 &9.45454545454545 \\
Llama-3.2-3B-Instruct &74.7967479674796 &0 &0 &95.4545454545454 &72.2415795586527 &76.2148337595907 &0 &1.13960113960113 &95.4545454545454 &72.2415795586527 \\
Llama-3.2-1B-Instruct &85.5539971949509 &76.6355140186915 &89.1963109354413 &85 &98.8970588235294 &85.3185595567867 &76.6355140186915 &89.1963109354413 &85 &98.8970588235294 \\
Qwen2-1.5B-Instruct &78.7775891341256 &64.3274853801169 &74.9442379182156 &96.045197740113 &70.4358068315665 &73.5177865612648 &61.5384615384615 &74.5304282494365 &96.045197740113 &70.4358068315665 \\
Qwen2.5-0.5B-Instruct &73.8522954091816 &66.6666666666666 &82.7057182705718 &98.342541436464 &89.4472361809045 &72.9166666666666 &66.2921348314606 &82.7057182705718 &98.342541436464 &89.4472361809045 \\
Qwen2.5-7B-Instruct-1M &5.55555555555555 &0 &0 &97.2067039106145 &0 &0 &0 &0 &97.2067039106145 &0 \\
Qwen2.5-14B-Instruct-1M &0 &0 &0 &87.8048780487805 &0 &0 &0 &0 &87.8048780487805 &0 \\

\midrule
\textbf{Baseline} & \multicolumn{10}{c}{\textsc{Oddballness}} \\ \cmidrule{2-11}
Llama-3.1-8B-Instruct &13.7931034482758 &0 &0 &73.103448275862 &2.24719101123595 &0 &0 &0 &74.8299319727891 &0 \\
Llama-3.2-3B-Instruct &64.8648648648648 &0 &0 &95.4545454545454 &72.2415795586527 &65.9793814432989 &4.25531914893617 &2.25352112676056 &95.4545454545454 &72.2415795586527 \\
Llama-3.2-1B-Instruct &85.2367688022284 &76.6355140186915 &89.1963109354413 &85 &98.8970588235294 &85.3185595567867 &76.6355140186915 &89.1963109354413 &85 &98.8970588235294 \\
Qwen2-1.5B-Instruct &77.6167471819645 &60.9756097560975 &74.9442379182156 &96.045197740113 &70.4358068315665 &73.9823008849557 &64.4295302013422 &74.7565543071161 &96.045197740113 &70.4358068315665 \\
Qwen2.5-0.5B-Instruct &73.6220472440944 &68.3937823834196 &82.7057182705718 &98.342541436464 &89.4472361809045 &72.0338983050847 &69.3877551020408 &82.7057182705718 &98.342541436464 &89.4472361809045 \\
Qwen2.5-7B-Instruct-1M &1.98019801980198 &0 &0 &97.2067039106145 &0 &0 &0 &0 &97.2067039106145 &0 \\
Qwen2.5-14B-Instruct-1M &0 &0 &0 &87.8048780487805 &0 &0 &0 &0 &87.8048780487805 &0 \\

\midrule
\textbf{Ours (Sentence-level)} & \multicolumn{10}{c}{\textsc{Surprisal + CIS + Entropy + CWS}} \\ \cmidrule{2-11}
Llama-3.1-8B-Instruct &27.7456647398843 &0 &0.714285714285714 &74.2857142857142 &31.6622691292876 &41.3223140495867 &0 &20.9424083769633 &66.6666666666666 &45.8752515090543 \\
Llama-3.2-3B-Instruct &74.8010610079575 &0 &15.0375939849624 &95.4545454545454 &71.9626168224299 &77.9746835443038 &27.3972602739726 &33.2737030411449 &95.4545454545454 &60.9609609609609 \\
Llama-3.2-1B-Instruct &85.6330014224751 &74.4897959183673 &89.0581717451523 &85 &98.8970588235294 &84.7507331378299 &71.1111111111111 &89.2409240924092 &78.6666666666666 &98.8970588235294 \\
Qwen2-1.5B-Instruct &77.3869346733668 &54.6762589928057 &71.3204373423044 &96.045197740113 &65.7824933687002 &73.0983302411873 &57.1428571428571 &70.3448275862068 &96.045197740113 &63.2503660322108 \\
Qwen2.5-0.5B-Instruct &75.3451676528599 &67.0807453416149 &81.9454679439941 &98.342541436464 &89.4472361809045 &75.4880694143167 &58.9743589743589 &81.83776022972 &97.7777777777777 &88.7080366225839 \\
Qwen2.5-7B-Instruct-1M &3.7037037037037 &0 &1.45985401459854 &97.2067039106145 &0 &26.9938650306748 &11.4285714285714 &23.5897435897435 &97.2067039106145 &8.40336134453781 \\
Qwen2.5-14B-Instruct-1M &0 &0 &0.78125 &87.8048780487805 &0 &5.71428571428571 &9.52380952380952 &26.0869565217391 &88.3435582822085 &33.5078534031413 \\

\midrule
\textbf{Ours} & \multicolumn{10}{c}{\textsc{Linguistically Targeted}} \\ \cmidrule{2-11}
Llama-3.1-8B-Instruct & {41.00 \textcolor{ForestGreen}{\scriptsize (+13.3)}} & {7.14 \textcolor{ForestGreen}{\scriptsize (+7.1)}} & {18.34 \textcolor{ForestGreen}{\scriptsize (+17.6)}} & {65.55 \textcolor{red}{\scriptsize (-8.7)}} & {48.02 \textcolor{ForestGreen}{\scriptsize (+16.4)}} & {50.00 \textcolor{ForestGreen}{\scriptsize (+8.7)}} & {10.53 \textcolor{ForestGreen}{\scriptsize (+10.5)}} & {42.07 \textcolor{ForestGreen}{\scriptsize (+21.1)}} & {65.45 \textcolor{red}{\scriptsize (-1.2)}} & {48.92 \textcolor{ForestGreen}{\scriptsize (+3.0)}} \\
Llama-3.2-3B-Instruct & {79.78 \textcolor{ForestGreen}{\scriptsize (+5.0)}} & {17.24 \textcolor{ForestGreen}{\scriptsize (+17.2)}} & {40.87 \textcolor{ForestGreen}{\scriptsize (+25.8)}} & {92.40 \textcolor{red}{\scriptsize (-3.1)}} & {65.79 \textcolor{red}{\scriptsize (-6.2)}} & {76.22 \textcolor{red}{\scriptsize (-1.8)}} & {32.56 \textcolor{ForestGreen}{\scriptsize (+5.2)}} & {49.40 \textcolor{ForestGreen}{\scriptsize (+16.1)}} & {94.86 \textcolor{red}{\scriptsize (-0.6)}} & {61.64 \textcolor{ForestGreen}{\scriptsize (+0.7)}} \\
Llama-3.2-1B-Instruct & {84.29 \textcolor{red}{\scriptsize (-1.3)}} & {75.86 \textcolor{ForestGreen}{\scriptsize (+1.4)}} & {89.18 \textcolor{ForestGreen}{\scriptsize (+0.1)}} & {77.85 \textcolor{red}{\scriptsize (-7.1)}} & \textcolor{gray}{\scriptsize (=)} & {81.90 \textcolor{red}{\scriptsize (-2.8)}} & {64.20 \textcolor{red}{\scriptsize (-6.9)}} & {84.55 \textcolor{red}{\scriptsize (-4.7)}} & {76.60 \textcolor{red}{\scriptsize (-2.1)}} & {98.80 \textcolor{red}{\scriptsize (-0.1)}} \\
Qwen2-1.5B-Instruct & {79.30 \textcolor{ForestGreen}{\scriptsize (+1.9)}} & {66.18 \textcolor{ForestGreen}{\scriptsize (+11.5)}} & {70.08 \textcolor{red}{\scriptsize (-1.2)}} & {96.00 \textcolor{red}{\scriptsize (-0.0)}} & {59.56 \textcolor{red}{\scriptsize (-6.2)}} & {74.24 \textcolor{ForestGreen}{\scriptsize (+1.1)}} & {63.45 \textcolor{ForestGreen}{\scriptsize (+6.3)}} & {70.12 \textcolor{red}{\scriptsize (-0.2)}} & \textcolor{gray}{\scriptsize (=)} & {57.96 \textcolor{red}{\scriptsize (-5.3)}} \\
Qwen2.5-0.5B-Instruct & {75.20 \textcolor{red}{\scriptsize (-0.1)}} & {63.69 \textcolor{red}{\scriptsize (-3.4)}} & {80.61 \textcolor{red}{\scriptsize (-1.3)}} & {96.05 \textcolor{red}{\scriptsize (-2.3)}} & {89.31 \textcolor{red}{\scriptsize (-0.1)}} & {71.55 \textcolor{red}{\scriptsize (-3.9)}} & {63.69 \textcolor{ForestGreen}{\scriptsize (+4.7)}} & {74.79 \textcolor{red}{\scriptsize (-7.0)}} & {98.34 \textcolor{ForestGreen}{\scriptsize (+0.6)}} & {83.60 \textcolor{red}{\scriptsize (-5.1)}} \\
Qwen2.5-7B-Instruct-1M & {26.85 \textcolor{ForestGreen}{\scriptsize (+23.1)}} & {22.86 \textcolor{ForestGreen}{\scriptsize (+22.9)}} & {9.80 \textcolor{ForestGreen}{\scriptsize (+8.3)}} & {97.75 \textcolor{ForestGreen}{\scriptsize (+0.5)}} & {24.55 \textcolor{ForestGreen}{\scriptsize (+24.5)}} & {42.62 \textcolor{ForestGreen}{\scriptsize (+15.6)}} & {12.77 \textcolor{ForestGreen}{\scriptsize (+1.3)}} & {39.55 \textcolor{ForestGreen}{\scriptsize (+16.0)}} & {97.18 \textcolor{red}{\scriptsize (-0.0)}} & {38.55 \textcolor{ForestGreen}{\scriptsize (+30.1)}} \\
Qwen2.5-14B-Instruct-1M & {9.88 \textcolor{ForestGreen}{\scriptsize (+9.9)}} & {11.11 \textcolor{ForestGreen}{\scriptsize (+11.1)}} & {13.94 \textcolor{ForestGreen}{\scriptsize (+13.2)}} & {83.87 \textcolor{red}{\scriptsize (-3.9)}} & {23.97 \textcolor{ForestGreen}{\scriptsize (+24.0)}} & {27.59 \textcolor{ForestGreen}{\scriptsize (+21.9)}} & {11.76 \textcolor{ForestGreen}{\scriptsize (+2.2)}} & {35.20 \textcolor{ForestGreen}{\scriptsize (+9.1)}} & {84.21 \textcolor{red}{\scriptsize (-4.1)}} & {40.38 \textcolor{ForestGreen}{\scriptsize (+6.9)}} \\

\bottomrule
\end{tabular}%
}

\caption{Results on logistic regression (left panel) and MLP (right panel) classifiers. Top three panels shows the results obtained using baseline features. Bottom two panels shows performance using features derived from surprisal cues, with and without the presence of linguistic spans. All values presented are F1 scores of detecting error, averaged across three runs.} 

\label{table:sentence_results}
\end{center}
\end{table*}

\vspace{-8pt}
\section{Results}
\label{sec:results}

\subsection{Can input likelihood features signal model failures?}

\Cref{table:sentence_results} reports F1 for error detection across models and datasets, comparing simple input-likelihood heuristics (mean log likelihood, mean max token probability, Oddballness) with our input-side feature sets.

For a given model–task pair, switching the classifier (LogReg vs. MLP) rarely changes outcomes for the baselines, and on the hardest benchmarks, DICE (idiomaticity) and TroFi/MOH-X (metaphor), they often show no signal (F1 = 0 for several large models, e.g., Llama-3.1-8B, Qwen2.5-7B, Qwen2.5-14B). This observation show that coarse confidence summaries are too blunt to capture the context-dependent failures these non-compositional tasks probe.

The sentence-level set (Surprisal + CIS + Entropy + CWS) yields substantial gains on every dataset and for every model family, turning many of the zero-F1 cases into non-trivial detection performance. The gains are especially visible on smaller models (0.5B–1.5B), where errors are more frequent but the input-conditioned belief patterns our features exploit are still systematic.

Classifier choice effects are modest but consistent. The MLP generally provides broader task coverage (fewer near-zero cases) and more stable performance across models and datasets. Logistic regression sometimes edges out MLP on smaller models ($\approx$0.5B–3B), which is consistent with linear separability of these features at lower capacity; the MLP tends to do better on the harder datasets (e.g., TroFi, ConMeC) and on larger models, indicating that our features encode non-trivial structure that benefits from a more expressive classifier.

\begin{table*}[h]
\begin{center}
\small
\resizebox{\linewidth}{!}{%
\begin{tabular}{lccccc|ccccc}
\toprule
Tasks & {DICE} & {MOHX} & {TroFi} & {PUB 14} & {ConMeC} & {DICE} & {MOHX} & {TroFi} & {PUB 14} & {ConMeC} \\ \midrule
\textbf{Ablation of CWS} & \multicolumn{10}{c}{\textsc{Surprisal + CIS + Entropy}} \\ \cmidrule{2-11}
Llama-3.1-8B-Instruct & \textcolor{red}{-1.3} & 0 & \textcolor{red}{-0.7} & \textcolor{ForestGreen}{2.0} & \textcolor{ForestGreen}{0.8} & \textcolor{red}{-2.2} & \textcolor{ForestGreen}{7.1} & \textcolor{red}{-1.8} & \textcolor{ForestGreen}{1.6} & \textcolor{red}{-10.4} \\
Llama-3.2-3B-Instruct & \textcolor{red}{-0.4} & 0 & \textcolor{red}{-6.7} & 0 & \textcolor{ForestGreen}{0.3} & \textcolor{ForestGreen}{0.5} & \textcolor{red}{-3.7} & \textcolor{ForestGreen}{3.7} & 0 & \textcolor{ForestGreen}{4.0} \\
Llama-3.2-1B-Instruct & \textcolor{ForestGreen}{0.1} & \textcolor{red}{-1.2} & \textcolor{ForestGreen}{0.1} & 0 & 0 & \textcolor{red}{-0.7} & \textcolor{red}{-1.7} & \textcolor{red}{-0.4} & \textcolor{red}{-1.4} & 0 \\
Qwen2-1.5B-Instruct & \textcolor{red}{-0.6} & \textcolor{ForestGreen}{8.1} & \textcolor{ForestGreen}{3.5} & 0 & \textcolor{ForestGreen}{3.8} & \textcolor{ForestGreen}{2.5} & \textcolor{red}{-3.0} & \textcolor{ForestGreen}{1.9} & 0 & \textcolor{red}{-8.0} \\
Qwen2.5-0.5B-Instruct & \textcolor{red}{-0.5} & \textcolor{red}{-3.5} & \textcolor{ForestGreen}{0.8} & 0 & 0 & \textcolor{red}{-1.5} & \textcolor{red}{-13.2} & \textcolor{red}{-0.1} & \textcolor{ForestGreen}{0.6} & \textcolor{ForestGreen}{0.3} \\
Qwen2.5-7B-Instruct-1M & \textcolor{ForestGreen}{3.4} & 0 & \textcolor{ForestGreen}{3.2} & 0 & 0 & \textcolor{red}{-14.2} & 0 & \textcolor{ForestGreen}{0.3} & 0 & \textcolor{ForestGreen}{10.7} \\
Qwen2.5-14B-Instruct-1M & 0 & 0 & \textcolor{red}{-0.0} & 0 & \textcolor{ForestGreen}{1.0} & \textcolor{ForestGreen}{5.2} & \textcolor{ForestGreen}{0.5} & \textcolor{red}{-3.9} & \textcolor{red}{-1.4} & \textcolor{red}{-8.7} \\\midrule

\textbf{Ablation of Surprisal} & \multicolumn{10}{c}{\textsc{CIS + Entropy + CWS}} \\ \cmidrule{2-11}
Llama-3.1-8B-Instruct & \textcolor{red}{-0.5} & 0 & \textcolor{red}{-0.7} & \textcolor{ForestGreen}{0.9} & \textcolor{ForestGreen}{0.3} & \textcolor{red}{-1.9} & \textcolor{ForestGreen}{7.1} & \textcolor{red}{-0.1} & 0 & \textcolor{red}{-4.6} \\
Llama-3.2-3B-Instruct & \textcolor{red}{-3.3} & 0 & \textcolor{red}{-5.3} & 0 & 0 & \textcolor{red}{-0.0} & \textcolor{red}{-4.0} & \textcolor{ForestGreen}{7.2} & 0 & \textcolor{ForestGreen}{5.8} \\
Llama-3.2-1B-Instruct & \textcolor{red}{-0.9} & \textcolor{red}{-1.5} & \textcolor{ForestGreen}{0.1} & 0 & 0 & \textcolor{red}{-0.5} & \textcolor{red}{-2.9} & \textcolor{red}{-0.4} & \textcolor{red}{-4.9} & 0 \\
Qwen2-1.5B-Instruct & \textcolor{red}{-0.6} & \textcolor{ForestGreen}{8.1} & \textcolor{ForestGreen}{3.5} & 0 & \textcolor{ForestGreen}{3.6} & \textcolor{ForestGreen}{1.5} & \textcolor{red}{-1.7} & \textcolor{ForestGreen}{2.6} & 0 & \textcolor{red}{-7.4} \\
Qwen2.5-0.5B-Instruct & \textcolor{red}{-0.7} & \textcolor{red}{-1.6} & \textcolor{ForestGreen}{0.8} & 0 & 0 & \textcolor{red}{-2.5} & \textcolor{red}{-7.2} & \textcolor{ForestGreen}{0.1} & \textcolor{ForestGreen}{0.6} & \textcolor{ForestGreen}{0.0} \\
Qwen2.5-7B-Instruct-1M & \textcolor{red}{-0.0} & 0 & \textcolor{ForestGreen}{1.2} & 0 & 0 & \textcolor{red}{-14.1} & 0 & \textcolor{ForestGreen}{2.6} & 0 & \textcolor{ForestGreen}{11.1} \\
Qwen2.5-14B-Instruct-1M & 0 & 0 & \textcolor{red}{-0.8} & 0 & \textcolor{ForestGreen}{1.0} & \textcolor{ForestGreen}{7.4} & \textcolor{ForestGreen}{0.5} & \textcolor{red}{-7.4} & \textcolor{red}{-2.1} & \textcolor{red}{-10.2} \\\midrule

\textbf{Ablation of CIS} & \multicolumn{10}{c}{\textsc{Surprisal + Entropy + CWS}} \\ \cmidrule{2-11}
Llama-3.1-8B-Instruct & \textcolor{red}{-0.6} & 0 & \textcolor{red}{-0.1} & \textcolor{red}{-2.5} & \textcolor{red}{-2.0} & \textcolor{red}{-1.1} & 0 & \textcolor{red}{-12.9} & \textcolor{ForestGreen}{7.2} & \textcolor{red}{-15.8} \\
Llama-3.2-3B-Instruct & \textcolor{ForestGreen}{0.8} & 0 & \textcolor{red}{-2.8} & 0 & \textcolor{ForestGreen}{0.3} & \textcolor{ForestGreen}{1.6} & \textcolor{red}{-6.2} & \textcolor{red}{-0.6} & 0 & \textcolor{ForestGreen}{5.1} \\
Llama-3.2-1B-Instruct & \textcolor{ForestGreen}{0.0} & \textcolor{red}{-1.4} & \textcolor{ForestGreen}{0.1} & 0 & 0 & \textcolor{red}{-0.3} & \textcolor{red}{-2.3} & \textcolor{red}{-0.1} & \textcolor{ForestGreen}{2.6} & 0 \\
Qwen2-1.5B-Instruct & \textcolor{red}{-0.7} & \textcolor{ForestGreen}{9.5} & \textcolor{ForestGreen}{3.4} & 0 & \textcolor{red}{-1.5} & \textcolor{ForestGreen}{1.1} & \textcolor{ForestGreen}{6.0} & \textcolor{ForestGreen}{3.2} & 0 & \textcolor{red}{-3.0} \\
Qwen2.5-0.5B-Instruct & \textcolor{red}{-0.6} & \textcolor{ForestGreen}{1.5} & \textcolor{ForestGreen}{0.8} & 0 & 0 & \textcolor{red}{-1.0} & \textcolor{red}{-5.6} & \textcolor{ForestGreen}{0.2} & \textcolor{ForestGreen}{0.6} & \textcolor{ForestGreen}{0.6} \\
Qwen2.5-7B-Instruct-1M & \textcolor{ForestGreen}{1.9} & 0 & \textcolor{ForestGreen}{0.5} & 0 & 0 & \textcolor{red}{-15.1} & \textcolor{red}{-11.4} & \textcolor{red}{-3.1} & 0 & \textcolor{ForestGreen}{6.7} \\
Qwen2.5-14B-Instruct-1M & 0 & 0 & \textcolor{red}{-0.8} & 0 & 0 & \textcolor{red}{-5.7} & \textcolor{red}{-9.5} & \textcolor{red}{-14.0} & \textcolor{red}{-0.7} & \textcolor{red}{-28.2} \\\midrule

\textbf{Ablation of Entropy} & \multicolumn{10}{c}{\textsc{Surprisal + CIS + CWS}} \\ \cmidrule{2-11}
Llama-3.1-8B-Instruct & \textcolor{red}{-0.2} & 0 & \textcolor{red}{-0.7} & \textcolor{ForestGreen}{0.2} & \textcolor{red}{-0.2} & \textcolor{red}{-1.3} & \textcolor{ForestGreen}{7.4} & \textcolor{red}{-5.7} & \textcolor{ForestGreen}{7.0} & \textcolor{red}{-16.9} \\
Llama-3.2-3B-Instruct & \textcolor{red}{-0.1} & 0 & \textcolor{red}{-6.9} & 0 & \textcolor{ForestGreen}{0.1} & \textcolor{red}{-0.6} & \textcolor{red}{-4.8} & \textcolor{red}{-4.5} & 0 & \textcolor{ForestGreen}{9.1} \\
Llama-3.2-1B-Instruct & \textcolor{red}{-0.3} & \textcolor{ForestGreen}{2.2} & \textcolor{ForestGreen}{0.1} & 0 & 0 & \textcolor{ForestGreen}{0.7} & \textcolor{ForestGreen}{1.1} & \textcolor{red}{-0.0} & \textcolor{ForestGreen}{0.2} & 0 \\
Qwen2-1.5B-Instruct & \textcolor{red}{-0.5} & \textcolor{ForestGreen}{3.4} & \textcolor{ForestGreen}{3.7} & 0 & \textcolor{ForestGreen}{0.3} & \textcolor{ForestGreen}{2.0} & \textcolor{red}{-1.1} & \textcolor{ForestGreen}{2.8} & 0 & \textcolor{red}{-2.9} \\
Qwen2.5-0.5B-Instruct & \textcolor{red}{-0.2} & \textcolor{red}{-0.4} & \textcolor{ForestGreen}{0.8} & 0 & 0 & \textcolor{red}{-3.3} & \textcolor{red}{-5.0} & \textcolor{ForestGreen}{0.1} & \textcolor{ForestGreen}{0.6} & \textcolor{ForestGreen}{0.5} \\
Qwen2.5-7B-Instruct-1M & \textcolor{ForestGreen}{1.8} & 0 & \textcolor{red}{-0.1} & 0 & 0 & \textcolor{red}{-25.0} & \textcolor{ForestGreen}{6.2} & \textcolor{red}{-16.7} & 0 & \textcolor{ForestGreen}{10.3} \\
Qwen2.5-14B-Instruct-1M & 0 & 0 & \textcolor{red}{-0.8} & 0 & 0 & \textcolor{red}{-5.7} & \textcolor{ForestGreen}{9.5} & \textcolor{red}{-19.3} & \textcolor{red}{-0.5} & \textcolor{red}{-20.5} \\
\bottomrule
\end{tabular}%
}
\caption{Feature ablation results for the logistic regression (left panel) and MLP (right panel) classifiers. Values represent the performance difference ($\Delta$) between the full results on all four metrics and the ablated model results, computed as: $\Delta = \text{ablated model} - \text{full model}$. All values are results averaged from three runs. \textcolor{red}{Negative values} indicate that removing the feature decreases performance (i.e., the feature is important), while \textcolor{ForestGreen}{{Positive} values} suggest the feature may be redundant or detrimental. Each panel shows the effect of ablating a specific metric.}

\label{table:ablation_results}
\end{center}
\end{table*}

\subsection{Do linguistically localized features improve error prediction?}

We next ask whether prediction improves when we localize input-side features to regions marked by the datasets as potential sites of misreading (e.g., the annotated phrase). 

As shown in \Cref{table:sentence_results}, span-localized features often provide stronger signal than sentence-level aggregation alone, particularly for larger models. For example, on Qwen2.5-14B (LogReg), localization yields clear improvements on ConMeC and TroFi, and Qwen2.5-7B shows consistent gains across multiple datasets. This suggests that larger models, while strong on average, benefit from focused measurement at the span and its boundary, where local instability can be more diagnostic than global summaries.

For smaller models (e.g., Llama-3.2-1B, Qwen2.5-0.5B), sentence-level features capture most of the available signal, and span-localization brings only modest, or occasionally negative, changes, consistent with errors that are predominantly global and already reflected in the input-likelihood profile.

Overall, sentence-level features provide a strong, pre-output signal on their own. Span-localized features are complementary, offering clear gains when region of interests are available (or can be inferred) and the base model has capacity for improvement. This points toward future work on automatic localization to bring these benefits without task-specific annotations.

\subsection{Where is the information for error?}
In this section, we aim to pinpoint the most important signals for error. To this end, we analyze the impact of removing individual surprisal-based cues from the sentence-level models by computing the performance delta across logistic regression and MLP classifiers. As shown in \Cref{table:ablation_results}, the ablations affect the MLP classifier more significantly, which reflects its higher sensitivity to input features and its capacity to model more complex, non-linear interactions than logistic regression.

For logistic regression, most changes are small (often within a few F1 points), with the larger negative deltas appearing when Surprisal or Entropy is removed. In contrast, CIS and CWS generally contribute less under the linear model, with ablations producing little movement in many model–task pairs.

The pattern shifts with the MLP: CIS and Entropy account for much of the signal, with their ablations producing the largest drops overall, while Surprisal remains useful but is no longer the dominant driver of performance. CWS shows the least consistent impact across both classifiers, since it is essentially surprisal augmented with a confidence/peakiness penalty and is therefore highly collinear with the explicit Surprisal and Entropy features already included, limiting its marginal contribution. Taken together, these results suggest that feature importance depends on the decision surface: a linear model primarily exploits Surprisal/Entropy, whereas a non-linear model leverages interactions that make CIS and Entropy comparatively more informative, with CWS contributing the least.

\section{Conclusions}
\label{sec:conclusions}

We showed that a language model’s \emph{perception of the input}, captured by its token-level probabilities over the given prompt, provides a reliable, \emph{pre-output} signal of error. Operationalizing this idea with simple, interpretable features over the input-side likelihood surface (Surprisal, Entropy, CIS, CWS), our framework anticipates failures without consulting generated outputs or hidden activations, requiring only token log-probabilities.
Across five context-sensitive benchmarks and a range of model scales, these structured input-side features consistently recover usable signal where coarse input-likelihood summaries (e.g., mean log likelihood, mean max token probability, Oddballness) often show little or no separability, especially on idiomaticity and metaphor. This establishes a clear result: {errors can be foretold from how the model reads the input, not just from what it later says.}
We further find a consistent deployment pattern in our experiments. Sentence-level features alone deliver strong performance, particularly for smaller models, while \emph{localizing} the measurement to annotated regions of interest yields substantial additional gains for larger models. In practice, this gives a simple recipe: use sentence-level features by default; add span-localized measurement when regions of interest are known or can be inferred.
This input-only view is complementary to output- or activation-based detectors and can be composed with them to further improve reliability.

\section*{Limitations}

While this work focuses on evaluation using English data, this work can be extended to other languages, given availability of evaluation data, as the relevant input signals can be generated and are expected to be similarly informative.      

Due to limited research budget, an important direction for future work involves extending our framework to evaluate and leverage cognitively inspired signals from closed-source models such as OpenAI's GPT-4o. These models are increasingly prevalent in deployed NLP systems, yet their opacity poses challenges for extracting internal metrics like layerwise activations or fine-tuned representations.

In our preliminary investigation, we also evaluated several smaller models to assess their performance. Namely, SmolLM models under 2B parameters \cite{smollm_paper}. However, we encountered a contradictory challenge: in order to test our classifiers, the LLM models must first be capable of generating meaningful responses to the non-literal language evaluation prompts. Unfortunately, the smaller models consistently failed to produce any correct outputs - they misclassified all instances. In other words, without at least some correct predictions to contrast against the errors, the requirement for error detection of distinguishing between right and wrong responses could not be met.

\section*{Acknowledgments}
This work was supported by the UKRI AI Centre for Doctoral Training in Speech and Language Technologies (SLT) and their Applications funded by UK Research and Innovation [grant number EP/S023062/1]. For the purpose of open access, the author has applied a Creative Commons Attribution (CC BY) licence to any Author Accepted Manuscript version arising. We acknowledge IT Services at The University of Sheffield for the provision of services for High Performance Computing. This research was partly supported by MRC-FAPESP AIM-Health and UniDive COST Action.
We are also greateful to members of the Sheffield NLP group for their helpful discussions in this work, in particular, Atsuki Yamaguchi, Ben Wu, Constantinos Karouzos. Finally, we thank the anonymous reviewers and metareviewer for their feedback on this paper.

\bibliography{anthology,custom}

\begin{thebibliography}{56}
\providecommand{\natexlab}[1]{#1}

\bibitem[{Allal et~al.(2025)Allal, Lozhkov, Bakouch, Blázquez, Penedo, Tunstall, Marafioti, Kydlíček, Lajarín, Srivastav, Lochner, Fahlgren, Nguyen, Fourrier, Burtenshaw, Larcher, Zhao, Zakka, Morlon, Raffel, von Werra, and Wolf}]{smollm_paper}
Loubna~Ben Allal, Anton Lozhkov, Elie Bakouch, Gabriel~Martín Blázquez, Guilherme Penedo, Lewis Tunstall, Andrés Marafioti, Hynek Kydlíček, Agustín~Piqueres Lajarín, Vaibhav Srivastav, Joshua Lochner, Caleb Fahlgren, Xuan-Son Nguyen, Clémentine Fourrier, Ben Burtenshaw, Hugo Larcher, Haojun Zhao, Cyril Zakka, Mathieu Morlon, and 3 others. 2025.
\newblock \href {https://arxiv.org/abs/2502.02737} {Smollm2: When smol goes big -- data-centric training of a small language model}.
\newblock \emph{Preprint}, arXiv:2502.02737.

\bibitem[{Ashok and May(2025)}]{ashok2025language}
Dhananjay Ashok and Jonathan May. 2025.
\newblock Language models can predict their own behavior.
\newblock \emph{arXiv preprint arXiv:2502.13329}.

\bibitem[{Azaria and Mitchell(2023)}]{azaria-mitchell-2023-internal}
Amos Azaria and Tom Mitchell. 2023.
\newblock \href {https://doi.org/10.18653/v1/2023.findings-emnlp.68} {The internal state of an {LLM} knows when it`s lying}.
\newblock In \emph{Findings of the Association for Computational Linguistics: EMNLP 2023}, pages 967--976, Singapore. Association for Computational Linguistics.

\bibitem[{Belrose et~al.(2023)Belrose, Furman, Smith, Halawi, Ostrovsky, McKinney, Biderman, and Steinhardt}]{tuned-lens}
Nora Belrose, Zach Furman, Logan Smith, Danny Halawi, Igor Ostrovsky, Lev McKinney, Stella Biderman, and Jacob Steinhardt. 2023.
\newblock \href {https://arxiv.org/abs/2303.08112} {Eliciting latent predictions from transformers with the tuned lens}.
\newblock \emph{Preprint}, arXiv:2303.08112.

\bibitem[{Birke and Sarkar(2006)}]{birke-sarkar-2006-clustering}
Julia Birke and Anoop Sarkar. 2006.
\newblock \href {https://aclanthology.org/E06-1042/} {A clustering approach for nearly unsupervised recognition of nonliteral language}.
\newblock In \emph{11th Conference of the {E}uropean Chapter of the Association for Computational Linguistics}, pages 329--336, Trento, Italy. Association for Computational Linguistics.

\bibitem[{Burns et~al.(2024)Burns, Ye, Klein, and Steinhardt}]{burns2024discoveringlatentknowledgelanguage}
Collin Burns, Haotian Ye, Dan Klein, and Jacob Steinhardt. 2024.
\newblock \href {https://arxiv.org/abs/2212.03827} {Discovering latent knowledge in language models without supervision}.
\newblock \emph{Preprint}, arXiv:2212.03827.

\bibitem[{Chafe(1968)}]{chafe1968idiomaticity}
Wallace~L Chafe. 1968.
\newblock Idiomaticity as an anomaly in the chomskyan paradigm.
\newblock \emph{Foundations of language}, pages 109--127.

\bibitem[{Chefer et~al.(2021)Chefer, Gur, and Wolf}]{Chefer_2021_CVPR}
Hila Chefer, Shir Gur, and Lior Wolf. 2021.
\newblock Transformer interpretability beyond attention visualization.
\newblock In \emph{Proceedings of the IEEE/CVF Conference on Computer Vision and Pattern Recognition (CVPR)}, pages 782--791.

\bibitem[{Chuang et~al.(2024)Chuang, Qiu, Hsieh, Krishna, Kim, and Glass}]{chuang-etal-2024-lookback}
Yung-Sung Chuang, Linlu Qiu, Cheng-Yu Hsieh, Ranjay Krishna, Yoon Kim, and James~R. Glass. 2024.
\newblock \href {https://doi.org/10.18653/v1/2024.emnlp-main.84} {Lookback lens: Detecting and mitigating contextual hallucinations in large language models using only attention maps}.
\newblock In \emph{Proceedings of the 2024 Conference on Empirical Methods in Natural Language Processing}, pages 1419--1436, Miami, Florida, USA. Association for Computational Linguistics.

\bibitem[{Church and Hanks(1990)}]{church-hanks-1990-word}
Kenneth~Ward Church and Patrick Hanks. 1990.
\newblock \href {https://aclanthology.org/J90-1003/} {Word association norms, mutual information, and lexicography}.
\newblock \emph{Computational Linguistics}, 16(1):22--29.

\bibitem[{Fano(1961)}]{fano_pmi}
R.M. Fano. 1961.
\newblock \href {https://books.google.ch/books?id=2PMbtQEACAAJ} {\emph{Transmission of Information: A Statistical Theory of Communication}}.
\newblock MIT Press Classics. MIT Press.

\bibitem[{Fraser(1970)}]{fraser1970idioms}
Bruce Fraser. 1970.
\newblock Idioms within a transformational grammar.
\newblock \emph{Foundations of language}, pages 22--42.

\bibitem[{Futrell et~al.(2020)Futrell, Gibson, and Levy}]{futrell_lossyContext_surprisal}
Richard Futrell, Edward Gibson, and Roger~P. Levy. 2020.
\newblock \href {https://doi.org/10.1111/cogs.12814} {Lossy-context surprisal: An information-theoretic model of memory effects in sentence processing}.
\newblock \emph{Cognitive Science}, 44(3):e12814.

\bibitem[{Genzel and Charniak(2002)}]{genzel-charniak-2002-entropy}
Dmitriy Genzel and Eugene Charniak. 2002.
\newblock \href {https://doi.org/10.3115/1073083.1073117} {Entropy rate constancy in text}.
\newblock In \emph{Proceedings of the 40th Annual Meeting of the Association for Computational Linguistics}, pages 199--206, Philadelphia, Pennsylvania, USA. Association for Computational Linguistics.

\bibitem[{Ghosh and Jiang(2025)}]{ghosh-jiang-2025-conmec}
Saptarshi Ghosh and Tianyu Jiang. 2025.
\newblock \href {https://aclanthology.org/2025.naacl-long.330/} {{C}on{M}e{C}: A dataset for metonymy resolution with common nouns}.
\newblock In \emph{Proceedings of the 2025 Conference of the Nations of the Americas Chapter of the Association for Computational Linguistics: Human Language Technologies (Volume 1: Long Papers)}, pages 6493--6509, Albuquerque, New Mexico. Association for Computational Linguistics.

\bibitem[{Goodkind and Bicknell(2018)}]{goodkind-bicknell-2018-predictive}
Adam Goodkind and Klinton Bicknell. 2018.
\newblock \href {https://doi.org/10.18653/v1/W18-0102} {Predictive power of word surprisal for reading times is a linear function of language model quality}.
\newblock In \emph{Proceedings of the 8th Workshop on Cognitive Modeling and Computational Linguistics ({CMCL} 2018)}, pages 10--18, Salt Lake City, Utah. Association for Computational Linguistics.

\bibitem[{Gralinski et~al.(2025)Gralinski, Staruch, and Jurkiewicz}]{gralinski-etal-2025-oddballness}
Filip Gralinski, Ryszard Staruch, and Krzysztof Jurkiewicz. 2025.
\newblock \href {https://aclanthology.org/2025.coling-main.183/} {Oddballness: universal anomaly detection with language models}.
\newblock In \emph{Proceedings of the 31st International Conference on Computational Linguistics}, pages 2683--2689, Abu Dhabi, UAE. Association for Computational Linguistics.

\bibitem[{Grattafiori et~al.(2024)Grattafiori, Dubey, Jauhri, Pandey, Kadian, Al-Dahle, Letman, Mathur, Schelten, Vaughan, Yang, Fan, Goyal, Hartshorn, Yang, Mitra, Sravankumar, Korenev, Hinsvark, Rao, Zhang, Rodriguez, Gregerson, Spataru, Roziere, Biron, Tang, Chern, Caucheteux, Nayak, Bi, Marra, McConnell, Keller, Touret, Wu, Wong, Ferrer, Nikolaidis, Allonsius, Song, Pintz, Livshits, Wyatt, Esiobu, Choudhary, Mahajan, Garcia-Olano, Perino, Hupkes, Lakomkin, AlBadawy, Lobanova, Dinan, Smith, Radenovic, Guzmán, Zhang, Synnaeve, Lee, Anderson, Thattai, Nail, Mialon, Pang, Cucurell, Nguyen, Korevaar, Xu, Touvron, Zarov, Ibarra, Kloumann, Misra, Evtimov, Zhang, Copet, Lee, Geffert, Vranes, Park, Mahadeokar, Shah, van~der Linde, Billock, Hong, Lee, Fu, Chi, Huang, Liu, Wang, Yu, Bitton, Spisak, Park, Rocca, Johnstun, Saxe, Jia, Alwala, Prasad, Upasani, Plawiak, Li, Heafield, Stone, El-Arini, Iyer, Malik, Chiu, Bhalla, Lakhotia, Rantala-Yeary, van~der Maaten, Chen, Tan, Jenkins, Martin, Madaan, Malo, Blecher,
  Landzaat, de~Oliveira, Muzzi, Pasupuleti, Singh, Paluri, Kardas, Tsimpoukelli, Oldham, Rita, Pavlova, Kambadur, Lewis, Si, Singh, Hassan, Goyal, Torabi, Bashlykov, Bogoychev, Chatterji, Zhang, Duchenne, Çelebi, Alrassy, Zhang, Li, Vasic, Weng, Bhargava, Dubal, Krishnan, Koura, Xu, He, Dong, Srinivasan, Ganapathy, Calderer, Cabral, Stojnic, Raileanu, Maheswari, Girdhar, Patel, Sauvestre, Polidoro, Sumbaly, Taylor, Silva, Hou, Wang, Hosseini, Chennabasappa, Singh, Bell, Kim, Edunov, Nie, Narang, Raparthy, Shen, Wan, Bhosale, Zhang, Vandenhende, Batra, Whitman, Sootla, Collot, Gururangan, Borodinsky, Herman, Fowler, Sheasha, Georgiou, Scialom, Speckbacher, Mihaylov, Xiao, Karn, Goswami, Gupta, Ramanathan, Kerkez, Gonguet, Do, Vogeti, Albiero, Petrovic, Chu, Xiong, Fu, Meers, Martinet, Wang, Wang, Tan, Xia, Xie, Jia, Wang, Goldschlag, Gaur, Babaei, Wen, Song, Zhang, Li, Mao, Coudert, Yan, Chen, Papakipos, Singh, Srivastava, Jain, Kelsey, Shajnfeld, Gangidi, Victoria, Goldstand, Menon, Sharma, Boesenberg,
  Baevski, Feinstein, Kallet, Sangani, Teo, Yunus, Lupu, Alvarado, Caples, Gu, Ho, Poulton, Ryan, Ramchandani, Dong, Franco, Goyal, Saraf, Chowdhury, Gabriel, Bharambe, Eisenman, Yazdan, James, Maurer, Leonhardi, Huang, Loyd, Paola, Paranjape, Liu, Wu, Ni, Hancock, Wasti, Spence, Stojkovic, Gamido, Montalvo, Parker, Burton, Mejia, Liu, Wang, Kim, Zhou, Hu, Chu, Cai, Tindal, Feichtenhofer, Gao, Civin, Beaty, Kreymer, Li, Adkins, Xu, Testuggine, David, Parikh, Liskovich, Foss, Wang, Le, Holland, Dowling, Jamil, Montgomery, Presani, Hahn, Wood, Le, Brinkman, Arcaute, Dunbar, Smothers, Sun, Kreuk, Tian, Kokkinos, Ozgenel, Caggioni, Kanayet, Seide, Florez, Schwarz, Badeer, Swee, Halpern, Herman, Sizov, Guangyi, Zhang, Lakshminarayanan, Inan, Shojanazeri, Zou, Wang, Zha, Habeeb, Rudolph, Suk, Aspegren, Goldman, Zhan, Damlaj, Molybog, Tufanov, Leontiadis, Veliche, Gat, Weissman, Geboski, Kohli, Lam, Asher, Gaya, Marcus, Tang, Chan, Zhen, Reizenstein, Teboul, Zhong, Jin, Yang, Cummings, Carvill, Shepard, McPhie,
  Torres, Ginsburg, Wang, Wu, U, Saxena, Khandelwal, Zand, Matosich, Veeraraghavan, Michelena, Li, Jagadeesh, Huang, Chawla, Huang, Chen, Garg, A, Silva, Bell, Zhang, Guo, Yu, Moshkovich, Wehrstedt, Khabsa, Avalani, Bhatt, Mankus, Hasson, Lennie, Reso, Groshev, Naumov, Lathi, Keneally, Liu, Seltzer, Valko, Restrepo, Patel, Vyatskov, Samvelyan, Clark, Macey, Wang, Hermoso, Metanat, Rastegari, Bansal, Santhanam, Parks, White, Bawa, Singhal, Egebo, Usunier, Mehta, Laptev, Dong, Cheng, Chernoguz, Hart, Salpekar, Kalinli, Kent, Parekh, Saab, Balaji, Rittner, Bontrager, Roux, Dollar, Zvyagina, Ratanchandani, Yuvraj, Liang, Alao, Rodriguez, Ayub, Murthy, Nayani, Mitra, Parthasarathy, Li, Hogan, Battey, Wang, Howes, Rinott, Mehta, Siby, Bondu, Datta, Chugh, Hunt, Dhillon, Sidorov, Pan, Mahajan, Verma, Yamamoto, Ramaswamy, Lindsay, Lindsay, Feng, Lin, Zha, Patil, Shankar, Zhang, Zhang, Wang, Agarwal, Sajuyigbe, Chintala, Max, Chen, Kehoe, Satterfield, Govindaprasad, Gupta, Deng, Cho, Virk, Subramanian, Choudhury,
  Goldman, Remez, Glaser, Best, Koehler, Robinson, Li, Zhang, Matthews, Chou, Shaked, Vontimitta, Ajayi, Montanez, Mohan, Kumar, Mangla, Ionescu, Poenaru, Mihailescu, Ivanov, Li, Wang, Jiang, Bouaziz, Constable, Tang, Wu, Wang, Wu, Gao, Kleinman, Chen, Hu, Jia, Qi, Li, Zhang, Zhang, Adi, Nam, Yu, Wang, Zhao, Hao, Qian, Li, He, Rait, DeVito, Rosnbrick, Wen, Yang, Zhao, and Ma}]{llama3_paper}
Aaron Grattafiori, Abhimanyu Dubey, Abhinav Jauhri, Abhinav Pandey, Abhishek Kadian, Ahmad Al-Dahle, Aiesha Letman, Akhil Mathur, Alan Schelten, Alex Vaughan, Amy Yang, Angela Fan, Anirudh Goyal, Anthony Hartshorn, Aobo Yang, Archi Mitra, Archie Sravankumar, Artem Korenev, Arthur Hinsvark, and 542 others. 2024.
\newblock \href {https://arxiv.org/abs/2407.21783} {The llama 3 herd of models}.
\newblock \emph{Preprint}, arXiv:2407.21783.

\bibitem[{Gu and Hopkins(2023)}]{gu-hopkins-2023-evaluation}
Zhengyao Gu and Mark Hopkins. 2023.
\newblock \href {https://doi.org/10.18653/v1/2023.acl-long.437} {On the evaluation of neural selective prediction methods for natural language processing}.
\newblock In \emph{Proceedings of the 61st Annual Meeting of the Association for Computational Linguistics (Volume 1: Long Papers)}, pages 7888--7899, Toronto, Canada. Association for Computational Linguistics.

\bibitem[{Hale(2001)}]{hale-2001-probabilistic}
John Hale. 2001.
\newblock \href {https://aclanthology.org/N01-1021/} {A probabilistic {E}arley parser as a psycholinguistic model}.
\newblock In \emph{Second Meeting of the North {A}merican Chapter of the Association for Computational Linguistics}.

\bibitem[{He et~al.(2024)He, Idiart, Scarton, and Villavicencio}]{he-etal-2024-enhancing}
Wei He, Marco Idiart, Carolina Scarton, and Aline Villavicencio. 2024.
\newblock \href {https://doi.org/10.18653/v1/2024.findings-acl.741} {Enhancing idiomatic representation in multiple languages via an adaptive contrastive triplet loss}.
\newblock In \emph{Findings of the Association for Computational Linguistics: ACL 2024}, pages 12473--12485, Bangkok, Thailand. Association for Computational Linguistics.

\bibitem[{Jaeger and Levy(2006)}]{jaegerLevy_2006}
T.~Jaeger and Roger Levy. 2006.
\newblock \href {https://proceedings.neurips.cc/paper_files/paper/2006/file/c6a01432c8138d46ba39957a8250e027-Paper.pdf} {Speakers optimize information density through syntactic reduction}.
\newblock In \emph{Advances in Neural Information Processing Systems}, volume~19. MIT Press.

\bibitem[{Jaeger(2010)}]{jaeger2010redundancy}
T~Florian Jaeger. 2010.
\newblock Redundancy and reduction: Speakers manage syntactic information density.
\newblock \emph{Cognitive psychology}, 61(1):23--62.

\bibitem[{Kabbara and Cheung(2022)}]{kabbara-cheung-2022-investigating}
Jad Kabbara and Jackie Chi~Kit Cheung. 2022.
\newblock \href {https://aclanthology.org/2022.coling-1.65/} {Investigating the performance of transformer-based {NLI} models on presuppositional inferences}.
\newblock In \emph{Proceedings of the 29th International Conference on Computational Linguistics}, pages 779--785, Gyeongju, Republic of Korea. International Committee on Computational Linguistics.

\bibitem[{Levy(2008)}]{levy-2008}
Roger Levy. 2008.
\newblock \href {https://doi.org/10.1016/j.cognition.2007.05.006} {Expectation-based syntactic comprehension}.
\newblock \emph{Cognition}, 106(3):1126--1177.

\bibitem[{Levy(2013)}]{levy2013}
Roger Levy. 2013.
\newblock Memory and surprisal in human sentence comprehension.
\newblock In \emph{Sentence processing.}, Current issues in the psychology of language., pages 78--114. Psychology Press, New York, NY, US.

\bibitem[{Li et~al.(2023)Li, Patel, Vi\'{e}gas, Pfister, and Wattenberg}]{ITI-li}
Kenneth Li, Oam Patel, Fernanda Vi\'{e}gas, Hanspeter Pfister, and Martin Wattenberg. 2023.
\newblock Inference-time intervention: eliciting truthful answers from a language model.
\newblock In \emph{Proceedings of the 37th International Conference on Neural Information Processing Systems}, NIPS '23, Red Hook, NY, USA. Curran Associates Inc.

\bibitem[{Ma et~al.(2025)Ma, Chen, Wang, and Zhang}]{ma2025estimating}
Huan Ma, Jingdong Chen, Guangyu Wang, and Changqing Zhang. 2025.
\newblock Estimating llm uncertainty with logits.
\newblock \emph{arXiv preprint arXiv:2502.00290}.

\bibitem[{McCoy et~al.(2024)McCoy, Yao, Friedman, Hardy, and Griffiths}]{mccoy2024embers}
R~Thomas McCoy, Shunyu Yao, Dan Friedman, Mathew~D Hardy, and Thomas~L Griffiths. 2024.
\newblock Embers of autoregression show how large language models are shaped by the problem they are trained to solve.
\newblock \emph{Proceedings of the National Academy of Sciences}, 121(41):e2322420121.

\bibitem[{Mi et~al.(2025)Mi, Villavicencio, and Moosavi}]{mi-etal-2025-rolling}
Maggie Mi, Aline Villavicencio, and Nafise~Sadat Moosavi. 2025.
\newblock \href {https://doi.org/10.18653/v1/2025.acl-long.362} {Rolling the {DICE} on idiomaticity: How {LLM}s fail to grasp context}.
\newblock In \emph{Proceedings of the 63rd Annual Meeting of the Association for Computational Linguistics (Volume 1: Long Papers)}, pages 7314--7332, Vienna, Austria. Association for Computational Linguistics.

\bibitem[{Mohammad et~al.(2016)Mohammad, Shutova, and Turney}]{mohammad-etal-2016-metaphor}
Saif Mohammad, Ekaterina Shutova, and Peter Turney. 2016.
\newblock \href {https://doi.org/10.18653/v1/S16-2003} {Metaphor as a medium for emotion: An empirical study}.
\newblock In \emph{Proceedings of the Fifth Joint Conference on Lexical and Computational Semantics}, pages 23--33, Berlin, Germany. Association for Computational Linguistics.

\bibitem[{Oh and Schuler(2023)}]{oh-schuler-2023-transformer}
Byung-Doh Oh and William Schuler. 2023.
\newblock \href {https://doi.org/10.18653/v1/2023.findings-emnlp.128} {Transformer-based language model surprisal predicts human reading times best with about two billion training tokens}.
\newblock In \emph{Findings of the Association for Computational Linguistics: EMNLP 2023}, pages 1915--1921, Singapore. Association for Computational Linguistics.

\bibitem[{Ohi et~al.(2024)Ohi, Kaneko, Koike, Loem, and Okazaki}]{ohi-etal-2024-likelihood}
Masanari Ohi, Masahiro Kaneko, Ryuto Koike, Mengsay Loem, and Naoaki Okazaki. 2024.
\newblock \href {https://doi.org/10.18653/v1/2024.findings-acl.193} {Likelihood-based mitigation of evaluation bias in large language models}.
\newblock In \emph{Findings of the Association for Computational Linguistics: ACL 2024}, pages 3237--3245, Bangkok, Thailand. Association for Computational Linguistics.

\bibitem[{Opedal et~al.(2024)Opedal, Chodroff, Cotterell, and Wilcox}]{opedal-etal-2024-role}
Andreas Opedal, Eleanor Chodroff, Ryan Cotterell, and Ethan Wilcox. 2024.
\newblock \href {https://doi.org/10.18653/v1/2024.emnlp-main.179} {On the role of context in reading time prediction}.
\newblock In \emph{Proceedings of the 2024 Conference on Empirical Methods in Natural Language Processing}, pages 3042--3058, Miami, Florida, USA. Association for Computational Linguistics.

\bibitem[{Pereyra et~al.(2017)Pereyra, Tucker, Chorowski, Łukasz Kaiser, and Hinton}]{pereyra2017regularizingneuralnetworkspenalizing}
Gabriel Pereyra, George Tucker, Jan Chorowski, Łukasz Kaiser, and Geoffrey Hinton. 2017.
\newblock \href {https://arxiv.org/abs/1701.06548} {Regularizing neural networks by penalizing confident output distributions}.
\newblock \emph{Preprint}, arXiv:1701.06548.

\bibitem[{Phelps et~al.(2024)Phelps, Pickard, Mi, Gow-Smith, and Villavicencio}]{phelps-etal-2024-sign}
Dylan Phelps, Thomas Pickard, Maggie Mi, Edward Gow-Smith, and Aline Villavicencio. 2024.
\newblock \href {https://aclanthology.org/2024.mwe-1.22/} {Sign of the times: Evaluating the use of large language models for idiomaticity detection}.
\newblock In \emph{Proceedings of the Joint Workshop on Multiword Expressions and Universal Dependencies (MWE-UD) @ LREC-COLING 2024}, pages 178--187, Torino, Italia. ELRA and ICCL.

\bibitem[{Pimentel et~al.(2023)Pimentel, Meister, Wilcox, Levy, and Cotterell}]{pimentel2013_readingTimes}
Tiago Pimentel, Clara Meister, Ethan~G. Wilcox, Roger~P. Levy, and Ryan Cotterell. 2023.
\newblock \href {https://doi.org/10.1162/tacl_a_00603} {On the effect of anticipation on reading times}.
\newblock \emph{Transactions of the Association for Computational Linguistics}, 11:1624--1642.

\bibitem[{Quevedo et~al.(2025)Quevedo, Salazar, Koerner, Rivas, and Cerny}]{quevedo}
Ernesto Quevedo, Jorge~Yero Salazar, Rachel Koerner, Pablo Rivas, and Tomas Cerny. 2025.
\newblock Detecting hallucinations in large language model generation: A token probability approach.
\newblock In \emph{Artificial Intelligence and Applications}, pages 154--173, Cham. Springer Nature Switzerland.

\bibitem[{Qwen et~al.(2025)Qwen, :, Yang, Yang, Zhang, Hui, Zheng, Yu, Li, Liu, Huang, Wei, Lin, Yang, Tu, Zhang, Yang, Yang, Zhou, Lin, Dang, Lu, Bao, Yang, Yu, Li, Xue, Zhang, Zhu, Men, Lin, Li, Tang, Xia, Ren, Ren, Fan, Su, Zhang, Wan, Liu, Cui, Zhang, and Qiu}]{qwen25_paper}
Qwen, :, An~Yang, Baosong Yang, Beichen Zhang, Binyuan Hui, Bo~Zheng, Bowen Yu, Chengyuan Li, Dayiheng Liu, Fei Huang, Haoran Wei, Huan Lin, Jian Yang, Jianhong Tu, Jianwei Zhang, Jianxin Yang, Jiaxi Yang, Jingren Zhou, and 25 others. 2025.
\newblock \href {https://arxiv.org/abs/2412.15115} {Qwen2.5 technical report}.
\newblock \emph{Preprint}, arXiv:2412.15115.

\bibitem[{Rambelli et~al.(2023)Rambelli, Chersoni, Senaldi, Blache, and Lenci}]{rambelli-etal-2023-frequent}
Giulia Rambelli, Emmanuele Chersoni, Marco S.~G. Senaldi, Philippe Blache, and Alessandro Lenci. 2023.
\newblock \href {https://doi.org/10.18653/v1/2023.mwe-1.13} {Are frequent phrases directly retrieved like idioms? an investigation with self-paced reading and language models}.
\newblock In \emph{Proceedings of the 19th Workshop on Multiword Expressions (MWE 2023)}, pages 87--98, Dubrovnik, Croatia. Association for Computational Linguistics.

\bibitem[{Shannon(1948)}]{shannon-1948}
C.~E. Shannon. 1948.
\newblock \href {https://doi.org/10.1002/j.1538-7305.1948.tb01338.x} {A mathematical theory of communication}.
\newblock \emph{The Bell System Technical Journal}, 27(3):379--423.

\bibitem[{Shorinwa et~al.(2024)Shorinwa, Mei, Lidard, Ren, and Majumdar}]{shorinwa2024survey}
Ola Shorinwa, Zhiting Mei, Justin Lidard, Allen~Z Ren, and Anirudha Majumdar. 2024.
\newblock A survey on uncertainty quantification of large language models: Taxonomy, open research challenges, and future directions.
\newblock \emph{arXiv preprint arXiv:2412.05563}.

\bibitem[{Smith and Levy(2013)}]{smith2013effect}
Nathaniel~J Smith and Roger Levy. 2013.
\newblock The effect of word predictability on reading time is logarithmic.
\newblock \emph{Cognition}, 128(3):302--319.

\bibitem[{Sravanthi et~al.(2024)Sravanthi, Doshi, Tankala, Murthy, Dabre, and Bhattacharyya}]{sravanthi-etal-2024-pub}
Settaluri Sravanthi, Meet Doshi, Pavan Tankala, Rudra Murthy, Raj Dabre, and Pushpak Bhattacharyya. 2024.
\newblock \href {https://doi.org/10.18653/v1/2024.findings-acl.719} {{PUB}: A pragmatics understanding benchmark for assessing {LLM}s' pragmatics capabilities}.
\newblock In \emph{Findings of the Association for Computational Linguistics: ACL 2024}, pages 12075--12097, Bangkok, Thailand. Association for Computational Linguistics.

\bibitem[{Steen et~al.(2010)Steen, Dorst, Herrmann, Kaal, Krennmayr, and Pasma}]{steen}
Gerard~J. Steen, Aletta~G. Dorst, J.~Berenike Herrmann, Anna~A. Kaal, Tina Krennmayr, and Tryntje Pasma. 2010.
\newblock \href {https://www.jbe-platform.com/content/books/9789027288158} {\emph{A Method for Linguistic Metaphor Identification}}.
\newblock John Benjamins.

\bibitem[{Tian et~al.(2024)Tian, Zhang, Xu, and Mao}]{tian-etal-2024-bridging}
Yuan Tian, Ruike Zhang, Nan Xu, and Wenji Mao. 2024.
\newblock \href {https://doi.org/10.18653/v1/2024.acl-long.719} {Bridging word-pair and token-level metaphor detection with explainable domain mining}.
\newblock In \emph{Proceedings of the 62nd Annual Meeting of the Association for Computational Linguistics (Volume 1: Long Papers)}, pages 13311--13325, Bangkok, Thailand. Association for Computational Linguistics.

\bibitem[{Tsipidi et~al.(2024)Tsipidi, Nowak, Cotterell, Wilcox, Giulianelli, and Warstadt}]{tsipidi-etal-2024-surprise}
Eleftheria Tsipidi, Franz Nowak, Ryan Cotterell, Ethan Wilcox, Mario Giulianelli, and Alex Warstadt. 2024.
\newblock \href {https://doi.org/10.18653/v1/2024.emnlp-main.1047} {Surprise! {U}niform {I}nformation {D}ensity isn`t the whole story: Predicting surprisal contours in long-form discourse}.
\newblock In \emph{Proceedings of the 2024 Conference on Empirical Methods in Natural Language Processing}, pages 18820--18836, Miami, Florida, USA. Association for Computational Linguistics.

\bibitem[{Vig and Belinkov(2019)}]{vig-belinkov-2019-analyzing}
Jesse Vig and Yonatan Belinkov. 2019.
\newblock \href {https://doi.org/10.18653/v1/W19-4808} {Analyzing the structure of attention in a transformer language model}.
\newblock In \emph{Proceedings of the 2019 ACL Workshop BlackboxNLP: Analyzing and Interpreting Neural Networks for NLP}, pages 63--76, Florence, Italy. Association for Computational Linguistics.

\bibitem[{Wilcox et~al.(2023)Wilcox, Pimentel, Meister, Cotterell, and Levy}]{wilcox_11_languages}
Ethan~G. Wilcox, Tiago Pimentel, Clara Meister, Ryan Cotterell, and Roger~P. Levy. 2023.
\newblock \href {https://doi.org/10.1162/tacl_a_00612} {Testing the predictions of surprisal theory in 11 languages}.
\newblock \emph{Transactions of the Association for Computational Linguistics}, 11:1451--1470.

\bibitem[{Wilks(1975)}]{WILKS1975}
Yorick Wilks. 1975.
\newblock \href {https://doi.org/10.1016/0004-3702(75)90016-8} {A preferential, pattern-seeking, semantics for natural language inference}.
\newblock \emph{Artificial Intelligence}, 6(1):53--74.

\bibitem[{Wilks(1978)}]{WILKS1978}
Yorick Wilks. 1978.
\newblock \href {https://doi.org/10.1016/0004-3702(78)90001-2} {Making preferences more active}.
\newblock \emph{Artificial Intelligence}, 11(3):197--223.

\bibitem[{Wolf et~al.(2020)Wolf, Debut, Sanh, Chaumond, Delangue, Moi, Cistac, Rault, Louf, Funtowicz, Davison, Shleifer, von Platen, Ma, Jernite, Plu, Xu, Scao, Gugger, Drame, Lhoest, and Rush}]{hf_paper}
Thomas Wolf, Lysandre Debut, Victor Sanh, Julien Chaumond, Clement Delangue, Anthony Moi, Pierric Cistac, Tim Rault, Rémi Louf, Morgan Funtowicz, Joe Davison, Sam Shleifer, Patrick von Platen, Clara Ma, Yacine Jernite, Julien Plu, Canwen Xu, Teven~Le Scao, Sylvain Gugger, and 3 others. 2020.
\newblock \href {https://arxiv.org/abs/1910.03771} {Huggingface's transformers: State-of-the-art natural language processing}.
\newblock \emph{Preprint}, arXiv:1910.03771.

\bibitem[{Wu et~al.(2025)Wu, Shen, Liu, Tang, Song, Wang, and Cai}]{wu-etal-2025-improve-decoding}
Jialiang Wu, Yi~Shen, Sijia Liu, Yi~Tang, Sen Song, Xiaoyi Wang, and Longjun Cai. 2025.
\newblock \href {https://aclanthology.org/2025.findings-naacl.217/} {Improve decoding factuality by token-wise cross layer entropy of large language models}.
\newblock In \emph{Findings of the Association for Computational Linguistics: NAACL 2025}, pages 3912--3921, Albuquerque, New Mexico. Association for Computational Linguistics.

\bibitem[{Xu and Reitter(2016)}]{xu-reitter-2016-entropy}
Yang Xu and David Reitter. 2016.
\newblock \href {https://doi.org/10.18653/v1/P16-1051} {Entropy converges between dialogue participants: Explanations from an information-theoretic perspective}.
\newblock In \emph{Proceedings of the 54th Annual Meeting of the Association for Computational Linguistics (Volume 1: Long Papers)}, pages 537--546, Berlin, Germany. Association for Computational Linguistics.

\bibitem[{Yang et~al.(2024)Yang, Chen, and Huang}]{yang-etal-2024-chatgpts}
Cheng Yang, Puli Chen, and Qingbao Huang. 2024.
\newblock \href {https://doi.org/10.18653/v1/2024.acl-long.57} {Can {C}hat{GPT}`s performance be improved on verb metaphor detection tasks? bootstrapping and combining tacit knowledge}.
\newblock In \emph{Proceedings of the 62nd Annual Meeting of the Association for Computational Linguistics (Volume 1: Long Papers)}, pages 1016--1027, Bangkok, Thailand. Association for Computational Linguistics.

\bibitem[{Zou et~al.(2025)Zou, Phan, Chen, Campbell, Guo, Ren, Pan, Yin, Mazeika, Dombrowski, Goel, Li, Byun, Wang, Mallen, Basart, Koyejo, Song, Fredrikson, Kolter, and Hendrycks}]{zou2025representationengineeringtopdownapproach}
Andy Zou, Long Phan, Sarah Chen, James Campbell, Phillip Guo, Richard Ren, Alexander Pan, Xuwang Yin, Mantas Mazeika, Ann-Kathrin Dombrowski, Shashwat Goel, Nathaniel Li, Michael~J. Byun, Zifan Wang, Alex Mallen, Steven Basart, Sanmi Koyejo, Dawn Song, Matt Fredrikson, and 2 others. 2025.
\newblock \href {https://arxiv.org/abs/2310.01405} {Representation engineering: A top-down approach to ai transparency}.
\newblock \emph{Preprint}, arXiv:2310.01405.

\end{thebibliography}

\appendix
\begin{table*}[ht]
\begin{center}
\small
\resizebox{\linewidth}{!}{%
\begin{tabular}{l *{5}{S} | *{5}{S}}
\toprule
Tasks & {DICE} & {MOHX} & {TroFi} & {PUB 14} & {ConMeC} & {DICE} & {MOHX} & {TroFi} & {PUB 14} & {ConMeC} \\ \midrule
Llama-3.1-8B-Instruct &20.6060606060606 &0 &0 &74.4827586206896 &28.8888888888888 &5.71428571428571 &0 &2.55591054313099 &66.6666666666666 &31.0626702997275 \\
Llama-3.2-3B-Instruct &74.5308310991957 &0 &0.564971751412429 &95.4545454545454 &72.3653395784543 &78.7128712871287 &15.625 &20.2643171806167 &95.4545454545454 &67.1052631578947 \\
Llama-3.2-1B-Instruct &85.1904090267983 &76.6355140186915 &89.1963109354413 &85 &98.8970588235294 &84.9507735583685 &72.7272727272727 &89.1963109354413 &84.2767295597484 &98.8970588235294 \\
Qwen2-1.5B-Instruct &79.7297297297297 &63.4730538922155 &74.9442379182156 &96.045197740113 &68.7338501291989 &76.4818355640535 &59.7402597402597 &74.0108611326609 &96.045197740113 &67.6240208877284 \\
Qwen2.5-0.5B-Instruct &73.558648111332 &65.2173913043478 &82.7057182705718 &98.342541436464 &89.4472361809045 &73.7068965517241 &60.6451612903225 &82.7972027972028 &98.342541436464 &89.4472361809045 \\
Qwen2.5-7B-Instruct-1M &3.73831775700934 &0 &3.33333333333333 &97.2067039106145 &0 &0 &0 &2.69360269360269 &97.2067039106145 &0 \\
Qwen2.5-14B-Instruct-1M &0 &0 &0 &87.8048780487805 &0 &0 &0 &1.47058823529411 &85.7142857142857 &1.85185185185185 \\
\bottomrule
\end{tabular}%
}
\caption{Results on logistic regression (left panel) and MLP (right panel) classifiers, trained on all three baselines. All values presented are F1 scores of detecting error, averaged across three runs.} 

\label{table:appendix_joined_baselines}
\end{center}
\end{table*}

\section{Additional Linguistics Features}
\label{sec:additional_features}

\paragraph{Context-level Span}
We define context-level span as considering the surprisal distribution, surround the targeted expression.
\[
    k_{\text{ctx}}(x) = k_{\text{left}}(x) \cup k_{\text{right}}(x),
\]
where:
\begin{align*}
    k_{\text{left}}(x) &\subset \{1, \dots, \min(k_{\text{expr}}(x)) - 1\}, \\
    k_{\text{right}}(x) &\subset \{\max(k_{\text{expr}}(x)) + 1, \dots, n\}.
\end{align*}

\paragraph{Relative magnitude of max surprisal in the idiom}

To assess the relative importance of surprisal within the idiom, we define the following ratio. Let:

\begin{equation}
s_{\max}^{\text{expr}} = \max_{i \in \mathcal{S}_{\text{expr}}} k(x_i), \quad S_{\text{rest}} = \sum_{i \in j \setminus \mathcal{S}_{\text{expr}}} k(x_i),
\end{equation}

\begin{equation}
R_{\text{contrast}} = \frac{s_{\max}^{\text{expr}}}{S_{\text{rest}} + \varepsilon},
\end{equation}

where \( \varepsilon > 0 \) is a small constant added for numerical stability. This feature quantifies how prominent the idiom’s most surprising token is relative to the rest of the sentence.

\paragraph{Position of min/max}

We find the position of the smallest and largest token in a sentence. Then, normalize this position by the total number of tokens in the sentence.

\[
i_{\min} = \arg\min_{1 \leq j \leq n} k(t_j), 
\quad
i_{\max} = \arg\max_{1 \leq j \leq n} k(t_j),
\]

\[
p_{\min} = \frac{i_{\min}}{n}, 
\quad 
p_{\max} = \frac{i_{\max}}{n},
\]

where \( k(t_j) \) is the scoring function applied to token \( t_j \).

\section{Experimental Setup}
\label{sec:experimental_setup}

\subsection{Prompts}
\label{sec:task_prompts}
\paragraph{DICE}
"Is the expression '\{target\_expression\}' used figuratively or literally in the sentence: \{sentence\} Answer 'i' for figurative, 'l' for literal.  Put your answer after 'output: '."
\paragraph{MOH-X and TroFi}
"Is the word '\{target\_word\}' used metaphorically or literally in the sentence: {sentence} Answer 'm' for metaphorical, 'l' for literal.  Put your answer after 'output: '."
\paragraph{PUB 14 Metonymy}
The PUB Task 14 dataset provides existing task instructions in this format:

\begin{quote}
\textbf{Context:} She is attracted to blue jacket. \\
\textbf{Question:} What does `blue jacket' refer to? \\
\textbf{Choices:} \[`Colour', `Jacket', `Sailor', `Sea'\]
\end{quote}

Therefore, we formulate our prompt as:

\begin{quote}
The following are multiple choice questions. \\
\texttt{[pretext]} \\
Your options are: \\
\texttt{[choices]}
\end{quote}

\paragraph{ConMeC}
"Is the word '\{target\_word\}' used metonymically or literally in the sentence: \{sentence\} Answer 'm' for metonymical, 'l' for literal.  Put your answer after 'output: '."

\subsection{Model Access}
We use HuggingFace \cite{hf_paper} to evaluate:
Llama 3 models \cite{llama3_paper} and Qwen 2.5 models \cite{qwen25_paper}.

We use two A100 GPUs to complete all our feature extraction aspects, as these require (1) prompting LLMs and (2) obtaining the logits from LLMs. 

The classifier part of our work can be run on CPUs.

\subsection{Classifiers}
\label{sec:hyperparameters}

\subsubsection{MLP} We employ a multi-layer perceptron to perform binary classification over $\featSet$. The architecture consists of three fully connected layers: an input layer mapping to 512 hidden units, a second hidden layer of 512 units, and a final output layer of a single unit. The two hidden layers are followed by ReLU activation functions, and the output layer employs a sigmoid activation to produce a scalar probability. This design follows that used in \cite{quevedo}.

Formally, given an input vector $x \in \mathbb{R}^d$, the output $y$ is computed as:

\[
\begin{aligned}
h_1 &= \text{ReLU}(W_1 x + b_1), \\
h_2 &= \text{ReLU}(W_2 h_1 + b_2), \\
y &= \sigma(W_3 h_2 + b_3),
\end{aligned}
\]
where $\sigma$ denotes the sigmoid function.

Prior to training, all features are standardized to zero mean and unit variance. 

We partition the dataset into training and validation splits using an 80/20 ratio, stratified to preserve class distribution across splits. This stratification is critical given the class imbalance commonly observed in error detection tasks.

Training is conducted using the binary cross-entropy loss, optimized via the Adam optimizer with a fixed learning rate of $10^{-3}$. Models are trained for 20 epochs with mini-batches of size 32. All computations are performed using PyTorch. At each epoch, we monitor both the binary cross-entropy loss and classification accuracy on the training set. Model evaluation is deferred until training completion.

\subsubsection{Logistic Regression}

We employ a regularized \textbf{logistic regression} classifier to perform binary classification over \errorLabelFull. Given an input vector $\featSet$, the predicted probability $y$ is given by:
\[
y = \sigma(w^\top x + b),
\]
where $w \in \mathbb{R}^d$ is the weight vector, $b$ is the bias term, and $\sigma(\cdot)$ denotes the sigmoid function.

We train the model using the \texttt{lbfgs} solver with an increased maximum number of iterations (2,500) to ensure convergence on the standardized feature set.
For each task, the dataset is split into training (80\%) and testing (20\%) sets using stratified sampling to maintain class balance across splits.

To ensure compatibility with linear models and to improve convergence, all features are standardized using \texttt{StandardScaler} to have zero mean and unit variance.

\subsection{Evaluation Metrics}

The performance of the classifiers is evaluated using the predicted labels on the held-out test set. We compute standard classification metrics including \textbf{precision}, \textbf{recall}, and \textbf{F1 score}, as well as class-wise scores. These metrics provide insights into the classifier’s ability to correctly identify both positive (error) and negative (non-error) instances.

\subsection{Baseline Definitions}
\label{sec:baselines_definitions}

\paragraph{Log Likelihood} 
In line with \cite{mccoy2024embers}, we compute log-likelihood of a text sequence using a causal language model.

\[
\frac{1}{n} \sum_{i=1}^{n} \log P(t_i \mid t_1, t_2, \dots, t_{i-1})
\]

\paragraph{Max Token Probability}
This function computes a confidence score for a given text based on a language model's predictions. For each token in the input (except the last, due to causal shifting), it calculates the maximum probability assigned to any token in the vocabulary at that position, which represents the model's confidence in its top prediction. These maximum probabilities are then aggregated using mean across the sequence using a specified method to produce a single scalar confidence value. This score reflects how confident the model is, on average or otherwise, about its predictions across the entire input sequence.
\[
\text{Confidence} = \frac{1}{n - 1} \sum_{i=1}^{n - 1} \max_{v \in V} P(v \mid t_1, \dots, t_{i})
\]

\paragraph{Oddballness}
We calculate the maximum Oddballness \cite{gralinski-etal-2025-oddballness} of a text sequence using a decoder-only language model. For each token in the sequence, it computes the softmax distribution over the vocabulary and compares the probability assigned to the actual token with all other tokens. The difference between these values (only when positive) is passed through a ReLU, summing the total surplus probability assigned to alternative tokens. This gives a per-token oddballness score, reflecting how much more likely the model thought other tokens were compared to the actual one. The function returns the maximum of these scores across the sequence, indicating the point where the model found the actual token most surprising or inconsistent.

\[
\text{oddball}_i = \sum_{v \in V} \text{ReLU}(P_i[v] - p_i)
\]

\section{Complementary Results}

\subsection{Model Performance on All Tasks}
\label{sec:downstream_accuracy}

\Cref{table:appendix_downstream_accuracy} presents the performance of the evaluated LLMs on all five non-literal language tasks.

\begin{table}[!htp]\centering
\resizebox{\columnwidth}{!}{%
\begin{tabular}{lrrrrrr}\toprule
Model &DICE &MOHX &TroFi &PUB 14 &ConMeC \\\midrule
Llama-3.1-8B-Instruct &69.8 &80.7 &62.8 &40.2 &54.0 \\
Llama-3.2-3B-Instruct &56.1 &66.8 &56.2 &8.7 &43.7 \\
Llama-3.2-1B-Instruct &25.7 &38.3 &20.9 &26.2 &2.2 \\
Qwen2-1.5B-Instruct &34.7 &48.8 &42.8 &7.2 &45.6 \\
Qwen2.5-0.5B-Instruct &43.5 &45.7 &31.2 &3.1 &18.9 \\
Qwen2.5-7B-Instruct-1M &76.3 &77.1 &63.5 &5.5 &62.4 \\
Qwen2.5-14B-Instruct-1M &83.7 &86.1 &66.9 &21.4 &62.3 \\
\bottomrule
\end{tabular}
}
\caption{Models performance results (classification accuracy) on each task.}
\label{table:appendix_downstream_accuracy}
\end{table}

\subsection{Combined Baselines}
We further evaluate a model using all three baseline feature sets jointly. The same training and evaluation protocols as in the main experiments are applied, except that the feature vectors are concatenated before training the classifiers. Results are presented in \Cref{table:appendix_joined_baselines}.

\end{document}